\documentclass{article}

\usepackage{arxiv}
\usepackage[utf8]{inputenc} % allow utf-8 input
\usepackage[T1]{fontenc}    % use 8-bit T1 fonts
\usepackage{hyperref}       % hyperlinks
\usepackage{url}            % simple URL typesetting
\usepackage{amsfonts}       % blackboard math symbols
\usepackage{nicefrac}       % compact symbols for 1/2, etc.
\usepackage{microtype}      % microtypography
\usepackage{lipsum}
\usepackage{amsmath}
\usepackage{amssymb}
\usepackage{amsthm}
\usepackage{graphicx}
\usepackage{glossaries, glossaries-extra}
\usepackage{xcolor}
\usepackage{soul}
\usepackage{hyperref}
\usepackage[ruled]{algorithm2e}
\usepackage{multirow}
\usepackage{makecell}
\usepackage{booktabs}
\usepackage{subfigure}

\graphicspath{ {./figures/} }

\newabbreviation{r2}{$R^2$}{coefficient of determination}
\newabbreviation{AS}{AS}{active subspace}
\newabbreviation{PDE}{PDE}{partial differential equation}
\newabbreviation{BGP}{B-GP}{Bayesian Gaussian process}
\newabbreviation{BAS}{B-FS}{Bayesian feature space}
\newabbreviation{MOAS}{MO-AS}{manifold optimization-based active subspace}
\newabbreviation{GP}{GP}{Gaussian process}
\newabbreviation{NUTS}{NUTS}{No-U-Turn Sampler}
\newabbreviation{MCMC}{MCMC}{Markov chain Monte-Carlo}
\newabbreviation{HMC}{HMC}{Hamiltonian Monte-Carlo}
\newabbreviation{BIC}{BIC}{Bayesian information criterion}
\newabbreviation{GPR}{GPR}{Gaussian process regression}
\newabbreviation{MC}{MC}{Monte-Carlo}
\newabbreviation{ARD}{ARD}{automatic relevance determination}
\newabbreviation{MLE}{MLE}{maximum likelihood estimation}
\newabbreviation{PCA}{PCA}{principal components analysis}
\newabbreviation{MLPPD}{MLPPD}{mean log pointwise predictive density}
\newabbreviation{TD}{TD}{training duration}
\newabbreviation{HDMR}{HDMR}{high-dimensional model representation}
\newabbreviation{RHMC}{RHMC}{Riemannian Hamiltonian Monte-Carlo}
\newabbreviation{SVD}{SVD}{singular value decomposition}
\newabbreviation{FS}{FS}{feature space}
\newabbreviation{FFD}{FFD}{free-form deformation}
\newabbreviation{MFSA}{MFSA}{mean first subspace angle}

\theoremstyle{definition}
\newtheorem{definition}{Definition}

\renewcommand{\vec}[1]{\mathbf{#1}}
\newcommand{\mat}[1]{\mathbf{#1}}
\newcommand{\rev}[1]{#1}

\title{A Fully Bayesian Gradient-Free Supervised Dimension Reduction Method using Gaussian Processes}

\usepackage{authblk}

\author[1\thanks{\tt{raphael.gautier@gatech.edu}}]{Raphaël Gautier}
\author[2]{Piyush Pandita}
\author[2]{Sayan Ghosh}
\author[1]{Dimitri Mavris}

\affil[1]{Aerospace Systems Design Laboratory, Georgia Institute of Technology, Atlanta, GA, 30332}
\affil[2]{Probabilistic Design, GE Research, Niskayuna, NY, 12309}

\begin{document}
\maketitle

% \Large
% \doublespacing

\begin{abstract}
Modern day engineering problems are ubiquitously characterized by sophisticated computer codes that map parameters or inputs to an underlying physical process. 
In other situations, experimental setups are used to model the physical process in a laboratory, ensuring high precision while being costly in materials and logistics.
In both scenarios, only limited amount of data can be generated by querying the expensive information source at a finite number of inputs or designs.
This problem is compounded further in the presence of a high-dimensional input space.
State-of-the-art parameter space dimension reduction methods, such as active subspace, aim to identify a subspace of the original input space that is sufficient to explain the output response.
These methods are restricted by their reliance on gradient evaluations or copious data, making them inadequate to expensive problems without direct access to gradients.
The proposed methodology is gradient-free and fully Bayesian, as it quantifies uncertainty in both the low-dimensional subspace and the surrogate model parameters. 
This enables a full quantification of epistemic uncertainty and robustness to limited data availability.
It is validated on multiple datasets from engineering and science and compared to two other state-of-the-art methods based on four aspects: a) recovery of the active subspace, b) deterministic prediction accuracy, c) probabilistic prediction accuracy, and d) training time.
The comparison shows that the proposed method improves the active subspace recovery and predictive accuracy, in both the deterministic and probabilistic sense, when only few model observations are available for training, at the cost of increased training time.

\end{abstract}


\glsdisablehyper

\section{Introduction}
\label{sec:intro}
\subsection{Motivation}
Engineering problems with high-dimensional inputs or parameters are often modeled in the form of expensive computer codes requiring significant computational resources and time, or as laboratory experiments incurring large labor and material costs.
These scenarios severely limit the number of data, namely input-output pairs, that can be observed.
This requires developing inexpensive-to-evaluate mathematical models that a) accurately model the underlying physical process and b) quantify the epistemic uncertainty due to limited observed data, where epistemic uncertainty is understood in the sense of Kiureghian and Ditlevsen~\cite{Kiureghian2009AleatoryMatter}.
Once developed, these data-driven probabilistic surrogate models can be used to learn arbitrary statistics about the output~\cite{oakley2004estimating}, generate more data intelligently~\cite{flournoy1993, Schonlau1997ComputerOptimization}, and make predictions under a limited budget~\cite{sacks1989}.
Adding to limited data availability, the curse of dimensionality remains a major obstacle in the way of developing such models when they admit a large number of input parameters.
The approach proposed in the present work aims at alleviating the impact of the curse of dimensionality on the creation of probabilistic surrogate models with high-dimensional inputs when observed data is limited.
\subsection{Previous Work}
While the present work focuses on the impact of the curse of dimensionality on the input space of surrogate models, it also affects their state and output spaces.
Addressing the high dimensionality of the state and output spaces has been the subject of extensive research in the fields of reduced-order modeling and equation-free model reduction~\cite{Benner2013ModelTools}. 
As a result, data-fit and physics-based surrogates of high-dimensional responses are now widespread in the literature. 
They leverage \emph{unsupervised} dimension reduction methods, such as \gls{PCA}~\cite{Jolliffe2005PrincipalScience}, deep autoencoder-decoder networks~\cite{Zhu2018BayesianQuantification}, or diffusion maps~\cite{Dietrich2018FastModel}. 
These methods are however not directly applicable to the problem of high-dimensional \emph{input} spaces, as they do not take the input-output relationship into account when learning the dimension reduction transformation. 
They may therefore lead to sub-optimal predictive performance because they discard information that is useful for predicting the response of interest. 
Reduction of the input space dimension demands a thorough understanding of the relationship between the inputs and the underlying function. 
As recognized in~\cite{Lataniotis2020ExtendingApproach}, \emph{supervised} techniques bring an advantage over unsupervised methods when applied in the context of surrogate modeling.
For some meta-modeling methods, such as Polynomial Chaos Expansion (PCE), an increase in input dimensionality may lead to a prohibitively high number of required model evaluations~\cite{Sargsyan2014DimensionalitySensing}. 
Among possible routes for alleviating the impact of the curse of dimensionality on the input space, the two deemed most promising in literature are: a) mapping strategies that take advantage of the dependence of the response on low-dimensional manifolds instead of the original high-dimensional input space, and b) lower-dimensional models, and in particular additive or partially additive models. While models in the latter category are motivated by many observations that high-order interactions are negligible in most scientific and engineering problems, e.g. \gls{HDMR}~\cite{Kubicek2015HighRepresentation}, the method proposed in this work belongs to the former category.

While mapping strategies are a broad denomination proposed in~\cite{Shan2010SurveyFunctions} in the context of engineering design, similar methods actually exist under other names in other fields:
approximation by ridge functions in the function approximation literature~\cite{Keiper2019ApproximationDimensions, Doerr2019TheDimensionality, Oglic2018ConstructiveAlgorithms, Glaws2018AIntegration, Kolleck2017OnSensing, Pinkus2015RidgeFunctions, DeVore1993ConstructiveApproximation, Tyagi2012LearningDimensions, Fornasier2012LearningDimensions, Cohen2012CapturingQueries, Schnass2011CompressedFunctions}, 
sufficient dimension reduction~\cite{Li2018SufficientR, Burges2009DimensionTour, Fukumizu2009KernelRegression, Adragni2009SufficientRegression, Li2007OnReduction}, 
spectral methods, sometimes referred to as supervised dimension reduction or dimension reduction for supervised learning~\cite{DeBie2005, Rosipal2006, Rosipal2011, Abdi2003, Yu2006SupervisedAnalysis, Chao2019RecentSurvey}, 
or probabilistic approaches~\cite{Tripathy2016, Tsilifis2018BayesianGeodesics}.
These methods aim at similar objectives while leveraging different mathematical tools. 
The term ridge function is usually used to refer to functions obtained by mapping the original high-dimensional space onto a low-dimensional subspace and extensive literature on this topic can be found. 
Recent methods were proposed to identify ridge functions, that rely on low-rank approximations~\cite{Fornasier2012LearningDimensions, Cohen2012CapturingQueries}.
Those methods can be interpreted as approximating finite-difference gradients, which is very ineffective and hinders exploration of the full input space by requiring many model evaluations around a limited number of points.
In the sufficient dimension reduction literature~\cite{Fukumizu2009KernelRegression, Ma2013AReduction}, methods such as sliced inverse regression~\cite{Ma2013AReduction} have been extensively used. 
However, as they assume elliptical distribution for the parameters, they do not fit the needs encountered when generating surrogate models, where the input space is usually uniformly sampled.
Hessian-based approaches have also been proposed to identify low-dimensional manifolds in the input space~\cite{Chen2019Hessian-basedReduction}.  
It is also interesting to note that while we are seeking methods that provide an explicit two-way mapping between the low- and high-dimensional spaces, the success of recent deep learning techniques can be attributed to an implicit dimensionality reduction, referred to as feature extraction~\cite{Guyon2008FeatureApplications}. 

The gradient-based \gls{AS} method~\cite{Constantine2014ActiveSurfaces} is among the most recent additions to the mapping or projection-based strategies. It has shown considerable promise on challenging engineering applications~\cite{Stoyanov2015AStochasticcoefficients, Constantine2015ExploitingScramjet, Constantine2016AcceleratingSubspaces, Tezzele2018DimensionProblems, Gross2020OptimisationMethod, Demo2020AProblems}. 
Extensions of the method to multivariate outputs~\cite{Zahm2020Gradient-basedFunctions, Rajaram2020Non-IntrusiveSubspace} and to the multi-fidelity setting~\cite{Lam2020MultifidelitySubspaces} are the subject of ongoing research.
Similar ideas were also independently pursued~\cite{Russi2010UncertaintyModels, Berguin2015MethodAnalyses}.
While the \gls{AS} method is in practice restricted to problems for which adjoint derivatives can be obtained, alternative formulations have been proposed to identify the low-dimensional structure of the input space using direct function evaluations only. 
To that end, some methods combine the projection onto the \gls{AS} with a probabilistic surrogate model such as \gls{GP} regression \cite{Tripathy2016, Seshadri2019DimensionFunctions, Rajaram2020Non-IntrusiveSubspace}. 
However, they only partially quantify epistemic uncertainty due to limited data, as the projection matrix on the \gls{AS} results from an optimization process instead of a Bayesian inference process. 
While similar approaches have been taken for other kinds of surrogates~\cite{Tipireddy2014BasisSpaces, Tsilifis2018BayesianGeodesics, Lataniotis2020ExtendingApproach}, the fully Bayesian inference of a predictive model combining a projection onto a lower-dimensional subspace and a Gaussian process has not yet been attempted.
Such an approach would open the door to efficient meta-modeling for high-dimensional problems using \gls{GP}s, including a full quantification of epistemic uncertainty introduced by limited data.

Including an orthonormal projection matrix in the predictive model complicates the Bayesian inference process, as readily available \gls{MCMC} samplers only operate with real-valued model parameters. 
To address this, it is useful to recognize that orthonormal matrices and linear subspaces are sets that can be equipped with a manifold structure, respectively forming the Stiefel and Grassmann manifolds.
The mathematical framework of manifolds grants a principled way of dealing with sets of relatively complex mathematical entities, such as orthonormal matrices, with the same tools that are routinely employed in Euclidean spaces, such as differentiation.
These manifolds are defined and their use in the contexts of optimization and Bayesian inference is discussed in section \ref{sec:manifolds}.
Methods to apply \gls{MCMC} when some model parameters belong to these manifolds have been developed following two distinct strategies: \gls{RHMC} techniques~\cite{Girolami2009RiemannianCarlo, Byrne2013GeodesicManifolds} and reparametrization techniques~\cite{Shepard2014TheFactors, Nirwan2019RotationPCA, Jauch2019MonteExpansion, Pourzanjani2017BayesianRepresentation}. 
The former modify the leapfrog steps of the original \gls{HMC} algorithm to ensure that new proposals remain on the relevant manifold. 
For the Stiefel manifold, this means that the orthogonality constraint is satisfied by construction.
The latter simply use parameterization techniques, such as Householder reflections~\cite{Nirwan2019RotationPCA}, polar angles~\cite{Jauch2019MonteExpansion}, the Givens representation~\cite{Pourzanjani2017BayesianRepresentation}, or simply a Gram-Schmidt orthogonalization scheme~\cite{Tripathy2018DeepQuantification} to transform a set of real-valued parameters into an orthogonal matrix.

While \gls{RHMC} methods are theoretically elegant and lie on strong mathematical foundations, they require the implementation of specialized \gls{MCMC} samplers~\cite{Betancourt2013GeneralizingManifolds}. 
“Turn-key” \gls{MCMC} algorithms that do not require extensive tuning have not yet been developed for \gls{RHMC}. 
As a consequence, implementations and usages of these methods are still at a very early stage of research.
The method recently proposed in \cite{Tsilifis2020BayesianProcesses} is representative of such approaches.
On the other hand, reparameterization approaches have been successfully leveraged using openly-available \gls{MCMC} samplers in instances where orthogonal matrices were part of the probabilistic model~\cite{Shepard2014TheFactors, Nirwan2019RotationPCA, Jauch2019MonteExpansion, Pourzanjani2017BayesianRepresentation}.
However, they have not yet been applied to the fully Bayesian inference of a low-dimensional input subspace in the context of supervised learning.
This is the approach taken in this work, eventually allowing us to leverage existing \gls{MCMC} algorithms to fully quantify uncertainty in the predictive model. By that, we mean both the uncertainty in the projection matrix onto a low-dimensional subspace as well as the uncertainty in the hyperparameters of the Gaussian process. In the following, because it can be thought of as a reduced set of relevant input features, we will refer to the low-dimensional linear subspace of the original input space that we are seeking as the \gls{FS}.
\subsection{Summary of Contributions}
Multiple contributions are made in this paper. The first contribution is a fully Bayesian and gradient-free formulation for meta-modeling of functions with high-dimensional inputs drawing inspiration from the \gls{AS} method. Under this formulation, all model parameters are considered uncertain, including those associated with the projection onto the lower-dimensional input subspace. The second contribution is a set of algorithms that enable the practical implementation of the proposed formulation using “turn-key” \gls{MCMC} samplers. Those additional implementation specifics, based on Householder transforms, are needed to simultaneously accommodate traditional \gls{GP} hyperparameters defined in Euclidean space and an orthonormal projection matrix belonging to the Stiefel manifold as model parameters during probabilistic inference. The third contribution is a thorough comparative study of the model’s performance with two recently proposed methods as seen from four different perspectives. The last contribution is a repository, written in the Python programming language and made openly available, that implements the proposed algorithms and benchmark methods using state-of-the-art probabilistic programming languages, manifold optimization libraries, and computational backends, enabling reproducibility and allowing interested researchers to develop extensions to the proposed method. 
\subsection{Paper Outline}
The outline of the rest of the paper is as follows. Our methodology is summarized in section \ref{sec:methodology}. In particular, subsection \ref{sec:fully_bayesian_as} introduces the method and algorithms proposed to parameterize the projection matrix and the fully Bayesian inference algorithm.
In section \ref{sec:results}, we demonstrate the performance of the methodology through numerical experiments on eight analytical functions of 25 to 100 inputs and four science and engineering datasets with input dimensions ranging from 18 to 100.
Since computationally expensive analyses highly constrain the number of observations available to train the surrogate model, the impact of the training set size on all metrics of interest is systematically assessed.
Finally, we summarize our findings in section \ref{sec:conclusion} and discuss natural extensions of the methodology to be explored next.

\newcommand{\inputsites}{\vec{x_1}, \dots, \vec{x_n}}
\section{Methodology}
\label{sec:methodology}
This section is organized as follows. Notation is first introduced in section \ref{sec:notation}. Background concepts and methods are then briefly summarized: Gaussian process regression in section \ref{sec:gpr}, the active subspace method in section \ref{sec:active_subspace}, Stiefel and Grassmann manifolds in section \ref{sec:manifolds}, and previous approaches to ridge approximation using Gaussian processes in section \ref{sec:gradient_free_as}. Finally, the proposed fully Bayesian approach to ridge approximation is detailed in section \ref{sec:fully_bayesian_as}.
\subsection{Notation}
\label{sec:notation}
Let $f$ be the mapping through which we collect data about the underlying physical process of interest. We assume that $f$ is a scalar-valued function of $d$ variables:
\begin{align}
\begin{split}
f: \mathcal{X} \subseteq \mathbb{R}^d &\longrightarrow \mathcal{Y} \subseteq \mathbb{R}\\
\vec{x} &\longmapsto y = f(\vec{x})
\end{split}
\end{align}
The input vector $\vec{x} \in \mathbb{R}^d$ while the output response $y \in \mathbb{R}$. The proposed approach targets high-dimensional input spaces, i.e. large values of $d$.

We assume that a total of $n$ input-output pairs, or model observations, have been obtained by evaluating the model $f$ at the input sites $\inputsites$ and are available to support the creation of the surrogate model. Following standard practice, we partition model observations into a training set $\mathcal{D}$ of size $p$ and a validation set $\mathcal{D}_*$ of size $q$ such that $p + q = n$. In the context of surrogate modeling, the observations in $\mathcal{D}$ are used for training the model while those in $\mathcal{D}_*$ allow to assess the performance of the predictive model.
\begin{align}
    \mathcal{D} &= \{(\vec{x_i}, y_i) \mid i = 1, \dots, p\} \\
    \mathcal{D}_* &= \{(\vec{x^*_i}, y^*_i) \mid i = 1, \dots, q\}
\end{align}
Let $\mat{X} = \left[ \vec{x_1}, \dots, \vec{x_p} \right] ^T$ and $\mat{X_*} = \left[ \vec{x^*_1}, \dots, \vec{x^*_q} \right] ^T$ respectively be the $p \times d$ and $q \times d$ design matrices corresponding to the training and validation input sites. Accordingly, let $\vec{y}$ and $\vec{y_*}$ be the size $p$ and $q$ training and validation response vectors. We focus on the creation of a surrogate model $\hat{f}$ of $f$, i.e. such that $\hat{f}$ can be evaluated in lieu of $f$.

We work within the probabilistic framework to quantify uncertainty in the surrogate model predictions. As such, we are seeking a generative model, i.e. a model for the joint probability distribution $p(\mat{X}, \vec{y})$. Probabilistic predictions for points in the validation set can then be made by using the conditional predictive distribution $p(\vec{y_*} \vert \mat{X_*}, \mat{X}, \vec{y})$. In the following, we denote a multivariate normal distribution with mean $\vec{m}$ and covariance matrix $\mat{V}$ as $\mathcal{N}(\vec{m}, \mat{V})$, and the identity matrix, whose size can be deduced from context, as $\mat{I}$.
\subsection{Gaussian Process Regression}
\label{sec:gpr}
\gls{GPR} is a popular surrogate modeling method that grants access to predictive uncertainty. 
We briefly recall the main idea behind \gls{GPR}, mostly adopting the reference notation introduced in \cite{williams2006gaussian}. 
The \gls{GPR} model relies on the assumption that the prior distribution for the underlying mapping $f$ can be modeled as a \gls{GP}: for a set of latent, unobserved, model responses made at input sites $\mat{X}$ and arranged in the vector $\vec{f}$, there exist a vector $\vec{\mu}$ and a matrix $\mat{\Sigma}$ such that $\vec{f} \sim \mathcal{N}(\vec{\mu}, \mat{\Sigma})$.

\rev{Additionally, it is assumed that observations $\vec{y}$ of $\vec{f}$ are independently affected by a zero-mean Gaussian random variable of variance $\sigma_n^2$ such that $\vec{y} \vert \vec{f} \sim \mathcal{N}(\vec{f}, \sigma_n^2 \mat{I})$.
This assumption is usually used to model the effect of measurement noise on experimental results. 
In the context of numerical simulations, measurement noise is irrelevant. 
However, the process of reducing the input space dimension by projecting it onto a low-dimensional subspace introduces artificial noise corresponding to the variations of the output due to variations of the inputs in the orthogonal complement of this low-dimensional subspace. 
In this work, this artificially introduced noise is accounted for by assuming noisy observations.}

Those assumptions enable the analytical marginalization of the latent vector $\vec{f}$, leading to the following prior distribution for the observations: $\vec{y} \sim \mathcal{N}(\vec{\mu}, \mat{\Sigma} + \sigma_n^2 \mat{I})$.
Multiple options for constructing the mean vector $\vec{\mu}$ and the covariance matrix $\mat{\Sigma}$ exist and lead to different \gls{GPR} variants. 
Common assumptions are made in this work. First, we assume the model observations to be centered and the mean of the prior \gls{GP} to be zero, i.e. $\vec{\mu} = [ 0, \dots, 0 ] ^T$. 
Then, a kernel function $k$ is used to construct $\mat{\Sigma}$ by encoding the correlation structure of the \gls{GP}, i.e. the correlation between responses $y$ and $y'$ at input input locations $\vec{x}$ and $\vec{x'}$. We use the widespread \gls{ARD} kernel:
\begin{equation}
    k(\vec{x}, \vec{x'}) = \sigma_f \prod_{i=1}^{d} \exp{\left( \frac{ \left( x_i-{x_i}' \right) ^2}{2 \ell_{i}^{2}} \right) }
\end{equation}
where $\sigma_f$ is the signal variance and $\vec{\ell} = \left[ \ell_{i}, ..., \ell_{d} \right]$ are characteristic length scales. For convenience, we denote as $K$ the generalization of the kernel function $k$ to design matrices. For two design matrices $\mat{X}$ and $\mat{X'}$ respectively containing $n$ and $n'$ points, we have:
\begin{equation}
    K(\mat{X}, \mat{X'}) = 
    \begin{bmatrix} 
        k(\vec{x_1}, \vec{x'_1}) & \dots & k(\vec{x_1}, \vec{x'_{n'}}) \\
        \vdots & \ddots & \vdots \\
        k(\vec{x_n}, \vec{x'_1}) & \dots & k(\vec{x_n}, \vec{x'_{n'}}) \\
    \end{bmatrix}   
\end{equation}
Using that notation, the GP covariance matrix is then computed as $\mat{\Sigma} = K(\mat{X}, \mat{X}).$
We gather all hyperparameters into the vector $\vec{\theta} = \left[ \sigma_n, \sigma_f, \ell_1, \dots, \ell_d \right]$. Rewriting the generative model with those assumptions and explicitly including parameters, we obtain equation \eqref{eq:gp_with_parameters}. Varying the hyperparameters $\vec{\theta}$ effectively leads to different generative models.
\begin{equation} \label{eq:gp_with_parameters}
    \vec{y} \vert \vec{\theta} \sim \mathcal{N}(0, K(\mat{X}, \mat{X};\sigma_f, \vec{\ell}) + \sigma_n^2 \mat{I})
\end{equation}
In a \gls{MLE} approach, training the \gls{GPR} model then consists in estimating the values of the hyperparameters $\vec{\theta}$ leading to the generative model that is most in agreement with the training data. This is achieved by selecting $\vec{\theta}$ that maximizes the likelihood $p(\vec{y} | \mat{X}, \vec{\theta})$. 
A closed-form equation for the likelihood $p(\vec{y} | \mat{X}, \vec{\theta})$ is made possible by the \gls{GP} assumption. 
In practice, the log-likelihood $\log{p(\vec{y}|X, \vec{\theta})}$ is used for numerical stability, Given training data $(\mat{X}, \vec{y})$ and denoting $\mat{K} = K(\mat{X}, \mat{X})$:
\begin{equation}
    \log{p(\vec{y}|X, \vec{\theta})} = -\frac{1}{2} \vec{y}^T \left( \mat{K} + \sigma_n^2 I \right) ^{-1} \vec{y} - \frac{1}{2} \log{\det{ \left( \mat{K} + \sigma_n^2 I \right) }} - \frac{n}{2} \log{2 \pi}
\end{equation}
In a Bayesian approach, hyperparameters are equipped with an prior distribution $p(\vec{\theta})$ and the full posterior distribution of the hyperparameters $p(\vec{\theta} | \vec{y}, \mat{X})$ is inferred by leveraging Bayes' rule:
\begin{equation}
    p(\vec{\theta} | \vec{y}, \mat{X}) \propto p(\vec{y} | \mat{X}, \vec{\theta}) p(\vec{\theta})
\end{equation}
Predictions $\vec{y_*}$ at validation locations $\mat{X_*}$ can be made by recalling that the underlying process is assumed to be a \gls{GP}, therefore the training and test outputs are distributed according to the following joint probability distribution:
\begin{equation}
    \begin{bmatrix} 
    \vec{y} \\ 
    \vec{y_*} 
    \end{bmatrix} 
    \sim
    \mathcal{N}
    \left(
        0, 
        \begin{bmatrix}
            K(\mat{X}, \mat{X}) + \sigma_n^2 \mat{I} & K(\mat{X}, \mat{X_*}) \\
            K(\mat{X_*}, \mat{X}) & K(\mat{X_*}, \mat{X_*}) + \sigma_n^2 \mat{I}
        \end{bmatrix} 
    \right)
\end{equation}
The posterior predictive distribution is obtained by conditioning the joint distribution with respect to the training data. Once again, the \gls{GP} assumption allows to derive this joint distribution analytically:
\begin{equation} \label{eq:posterior_predictive}
    \vec{y_*} | \mat{X_*}, \mat{X}, \vec{y}, \vec{\theta}
    \sim
    \mathcal{N}(\vec{\mu_*}, \mat{\Sigma_*})
\end{equation}
with:
\begin{align} 
    \vec{\mu_*} &= K(\mat{X_*}, \mat{X}) \left( K(\mat{X}, \mat{X}) + \sigma_n^2 I \right)^{-1} \vec{y} \label{eq:mu_star}\\
    \mat{\Sigma_*} &= K(\mat{X_*}, \mat{X_*}) - K(\mat{X_*}, \mat{X}) \left( K(\mat{X}, \mat{X})+\sigma_n^2 I \right) ^{-1} K(\mat{X}, \mat{X_*}) \label{eq:Sigma_star}
\end{align}
In the fully Bayesian approach to \rev{\gls{GPR}}, the hyperparameters need to be marginalized out using their posterior distribution to make predictions:
\begin{equation} \label{eq:bayesian_gp_predictive}
    p(\vec{y_*} | \mat{X_*}, \mat{X}, \vec{y}) = \int{p(\vec{y_*} | \mat{X_*}, \mat{X}, \vec{y}, \vec{\theta}) p(\vec{\theta} | \mat{X}, \vec{y}) \, d{\vec{\theta}}}
\end{equation}
\subsection{Active Subspace}
\label{sec:active_subspace}
The \glsxtrfull{AS} method addresses challenges raised by functions of high-dimensional inputs and the resulting curse of dimensionality that hinders the use of such functions in numerical activities such as uncertainty quantification, surrogate modeling, or numerical optimization.
In simple terms, the method seeks directions in input space that contribute in average the most to the variation of the output.
Those directions form the basis of a low-dimensional subspace of the original input space, the so-called \emph{active subspace}.
The curse of dimensionality is alleviated by substituting the active subspace to the original high-dimensional input space, obtaining a new, approximate mapping whose input space is effectively lower than the original.
\rev{It is} a general-purpose method in the sense this alternate mapping can be used to assist any of the aforementioned numerical applications. 

\rev{The \gls{AS} method comes within the general scope of \emph{approximation by ridge functions} ~\cite{Pinkus1997ApproximatingFunctions} that seek to approximate the function of interest $f$ with a mapping of the sort $f(\vec{x}) = g(\mat{W}^T \vec{x})$ where $\mat{W}$ is a tall projection matrix onto a subspace of the input space $\mathcal{X}$ and $g$ is a mapping whose input space is thus lower-dimensional than $f$'s.
While the \gls{AS} method does not necessarily yield the subspace leading to the optimal ridge approximation \cite{Constantine2017AApproximation}, where optimality is defined with respect to the squared prediction error, it has been shown to result in useful ridge approximation models for numerous engineering applications \cite{Stoyanov2015AStochasticcoefficients, Constantine2015ExploitingScramjet, Constantine2016AcceleratingSubspaces, Tezzele2018DimensionProblems, Gross2020OptimisationMethod, Demo2020AProblems}.}

The following paragraphs recall the major results pertaining to \rev{the \glsxtrshort{AS} method}. Interested readers \rev{may} find more details in \cite{Constantine2015ActiveStudies}.
In the context of \rev{this} method, we equip the input variables $\vec{x}$ of $f$ with a probability distribution $p(\vec{x})$.
\rev{The matrix $\mat{C}$, which is average of the outer product of the gradient with itself,}  plays a central role in the \gls{AS} method:
\begin{equation}
    \label{eq:gradient_covariance_matrix}
    \mat{C} = \int (\nabla_{\vec{x}} f) (\nabla_{\vec{x}} f)^T  p(\vec{x}) \, d\vec{x}
\end{equation}
\rev{This real symmetric matrix} is diagonalized as $\mat{C} = \mat{Q} \mat{\Lambda} \mat{Q}^T$ where $\mat{Q} = \left[ \vec{w_1}, \dots, \vec{w_d} \right]$ is the matrix containing the normalized eigenvectors $\left\{ \vec{w_i} \mid i=1,\dots,d \right\}$ of $\mat{C}$ and $\mat{\Lambda}$ is a diagonal matrix whose diagonal contains the eigenvalues $\left\{ \lambda_i \mid i=1,\dots,d \right\}$ of $\mat{C}$. We assume the eigenvalues to be sorted such that $\lambda_1 \geq \dots \geq \lambda_d \geq 0$.

The following relationship links the eigenvalues with the projections of the function's gradient onto the corresponding eigenvectors \cite{Constantine2015ActiveStudies}. For $i = 1, \dots, d$:
\begin{equation}
    \lambda_i = \int \left( \left( \nabla_{\vec{x}} f \right) ^ T \vec{w_i} \right)^2  p(\vec{x}) \, d\vec{x}
\end{equation}
The higher the $\lambda_i$, the greater the variations in $f$ in the direction \rev{$\vec{w_i}$}. The active subspace is defined as the subspace spanning the first $m \leq d$ directions $\left\{ \vec{w_i} \mid i=1,\dots,m \right\}$. In this subspace, the variations of $f$ are in average greater than in its orthogonal complement, referred to as the \emph{inactive subspace}.

Because they are the eigenvectors of a real symmetric matrix, the vectors $\left\{ \vec{w_i} \mid i=1,\dots,d \right\}$ form an orthonormal basis of the input space. We can then arrange them into two matrices:
\begin{align}
    \mat{W}  &= \left[ \vec{w_1}, \dots, \vec{w_m} \right] \\
    \mat{W_i}  &= \left[ \vec{w_{m+1}}, \dots , \vec{w_d} \right] 
\end{align}
$\mat{W}$ is the $d \times m$ projection matrix onto the \emph{active} subspace while $\mat{W_i}$ is the $d \times (d-m)$ projection matrix onto the \emph{inactive} subspace.
The original mapping of interest can then be rewritten as $f(\vec{x}) = f(\mat{W} \vec{z} + \mat{W_i} \vec{z_i})$ where $\vec{z} = \mat{W}^T \vec{x}$ is the component of the inputs in the AS and $\vec{z_i} = \mat{W_i}^T \vec{x}$ is the component of the inputs in the inactive subspace. Starting from this decomposition, a series of approximations can be made to obtain a practical AS-assisted surrogate modeling approach. 
Given the conditional probability $p(\vec{z_i}|\vec{z})$ of the inactive variables given the active variable, we start by defining the \emph{link function} $g$ as the conditional expectation of $f$ over the inactive subspace given a position in the active subspace:
\begin{equation}
    g(\vec{z}) = \mathbb{E}_{\vec{z_i} \vert \vec{z}} \left[  f(\mat{W} \vec{z} + \mat{W_i} \vec{z_i}) \right]
\end{equation}
By assuming that the variations of $f$ caused by variations of input variables in the inactive subspace are substantially less than those caused by variations in the inactive subspace, we obtain the following approximation:
\begin{equation} \label{eq:average_out_is}
    f(\vec{x}) \approx g(\mat{W}^T \vec{x})
\end{equation}
Since the exact integration in \eqref{eq:average_out_is} would be either too costly or simply not possible, a Monte-Carlo approximation $\hat{g}$ of $g$ is used:
\begin{equation}  \label{eq:link_function_mc}
    f(\vec{x}) \approx \hat{g}(\mat{W}^T \vec{x})
\end{equation}
The Monte-Carlo integration is shown to have good convergence properties in \cite{Constantine2015ActiveStudies} since $f$ does not by construction greatly vary in the inactive subspace. The function $\hat{g}$ is itself approximated by a surrogate model $\tilde{g}$ based on a limited number of model observations:
\begin{equation} \label{eq:link_function_data_fit}
    f(\vec{x}) \approx \tilde{g} (\mat{W}^T \vec{x}) 
\end{equation}
Finally, an approximation $\hat{\mat{W}}$ of $\mat{W}$ is obtained by replacing the integral with a finite sum in the computation of $\mat{C}$ in equation \eqref{eq:gradient_covariance_matrix}:
\begin{equation} \label{eq:mc_covariance_matrix}
    f(\vec{x}) \approx \tilde{g} (\hat{\mat{W}}^T \vec{x}) 
\end{equation}
AS-assisted surrogate modeling methods rely on equation \eqref{eq:mc_covariance_matrix} to approximate the original function $f$ \cite{Constantine2015ActiveStudies} by following a two-step approach. An approximation $\hat{\mat{W}}$ of the projection matrix $\mat{W}$ onto the AS is first computed using a sample of gradient evaluations. 
An approximation $\tilde{g}$ of the link function $g$ is then constructed using traditional surrogate modeling techniques \rev{by substituting the projected design matrix $\mat{Z} = \mat{X} \mat{W}$ to the original design matrix $\mat{X}$ in the training process}. 
\subsection{Stiefel and Grassmann Manifolds}
\label{sec:manifolds}
\rev{The previous section highlighted the central role played by orthogonal projection matrices in the \gls{AS} method and more generally in approximations by ridge functions.
The presence of an orthonormal projection matrix in the predictive model may be dealt with using real-valued model parameters and additional sets of constraints.
However, satisfying the orthonormality constraint adds to the computational burden of training the predictive model.
Instead, we recognize that the set of orthonormal matrices may be equipped with a manifold structure. 
A detailed explanation of the mathematical concepts surrounding manifolds is out of the scope of this paper, and a clear introduction to those can be found in \cite{Absil2009OptimizationManifolds}, from which we adapted the definitions given in this section.
A main benefit of working in manifolds is the well-defined transposition of differential calculus operations routinely made with real-valued parameters to orthonormal matrices.}

\rev{The set of orthonormal matrices forms the Stiefel manifold~\cite{Absil2009OptimizationManifolds}, whose definition, adapted from ~\cite{Absil2009OptimizationManifolds}, is given below.
\begin{definition}[Stiefel Manifold]
Let $\mathrm{St}(p, n)$ ($p \le n$) denote the set of all $n$ × $p$ orthonormal matrices 
\begin{equation}
    \{\mat{X} \in \mathbb{R}^{n \times p} : \mat{X}^T \mat{X} = \mat{I_p}\}
\end{equation}
where $I_p$  denotes the $p \times p$ identity matrix. 
Endowed with its manifold structure, the set $\mathrm{St}(p, n)$ is called the Stiefel manifold.
\end{definition}
A related manifold is the Grassmann manifold, defined below.
\begin{definition}[Grassmann Manifold]
Let $\mathrm{Gr}(p, n)$ be the set of all $p$-dimensional subspaces of $\mathbb{R}^n$. Endowed with its manifold structure, the set $\mathrm{Gr}(p, n)$ is called the Grassmann manifold
\end{definition}
By these definitions, given $p$ and $n$, we note that for every element $g \in \mathrm{Gr}(p, n)$ of the Grassmann manifold, we may find infinitely many elements $\{s \in \mathrm{St}(p, n): \mathrm{span}(s) = g\}$ in the Stiefel manifold whose span is $g$, all being orthonormal bases of the subspace $g$ that only differ by a rotation within $g$. As such, going from a orthornormal projection matrix in the Stiefel manifold to a subspace in the Grassmann manifold may be interpreted as retaining the information regarding the subspace spanned by this projection matrix, but losing the information regarding the exact orientation of the coordinate axes described by the matrix within that subspace.}
\subsection{Gaussian Processes with Built-In Dimensionality Reduction}
\label{sec:gradient_free_as}
\rev{When direct or cheap gradient evaluations are not available, the \gls{AS} method presented in section \ref{sec:active_subspace} may not be practically applicable, as gradient values would first need to be estimated, e.g. using finite differences, and such schemes require a large number of direct function evaluations when the input dimension $d$ is high.}

An alternative approach proposed in \cite{Tripathy2016} is to simultaneously train the approximate link function $\tilde{g}$ and the approximate projection matrix $\mat{\hat{W}}$ onto the AS.
The underlying predictive model combines aspects of the original \glsxtrshort{AS} method with Gaussian processes: the original inputs are projected onto a low-dimensional subspace that serves as the alternate, low-dimensional input space for a GP. This leads to the following \rev{generative} model:
\begin{equation} \label{eq:generative_model_with_projection}
    \vec{y} | \vec{\theta}, \mat{W} 
    \sim
    \mathcal{N} \left(
        0, 
        K \left( \mat{X} \mat{W}, \mat{X} \mat{W}; \vec{\theta} \right) \rev{+ \sigma_n^2 \mat{I}}
    \right)    
\end{equation}

This model can be broken down into two steps. 
An initial \emph{projection step} where the original design matrix $\mat{X}$ is projected in the AS to obtain a lower-dimensional design matrix $\mat{Z} = \mat{X} \mat{W}$ followed by a \emph{regression step} in which the lower-dimensional space is substituted to the original high-dimensional input space.

Compared to traditional GPR, this approach leads to the effective dimension reduction of GP's input space from the original $d$ inputs to only $m$ inputs. 
As a consequence, the number of GP hyperparameters is also decreased from $d + 2$ to $m + 2$ \rev{when using the \gls{ARD} kernel}. 
However, the projection matrix $\mat{W}$ is introduced as a new model parameter that must be determined in the training process.
\rev{Previous methods have leveraged the manifolds introduced in section \ref{sec:manifolds} to handle the projection matrix, which is taken as an element of the Stiefel manifold in~\cite{Tripathy2016, Seshadri2019DimensionFunctions} or Grassmann manifold in ~\cite{Rajaram2020Non-IntrusiveSubspace}. 
While they have distinct specifics, these methods all rely on optimization algorithms in manifolds, for which theory is well-established~\cite{Absil2009OptimizationManifolds} and numerical implementations are readily available, such as \emph{Pymanopt}~\cite{Townsend2016Pymanopt:Differentiation} used in~\cite{Rajaram2020Non-IntrusiveSubspace}}.
\subsection{Proposed Fully Bayesian Approach}
\label{sec:fully_bayesian_as}
\rev{The optimization-based approaches discussed in section \ref{sec:gradient_free_as}} do not enable a full quantification of epistemic uncertainty due to limited model observations. 
After projection of the design matrix onto the \rev{\glsxtrlong{FS}}, predictive uncertainty only originates from the assumption that the link function is modeled as a GP. 
However, neither the uncertainty in the GP hyperparameters nor in the projection matrix $\mat{W}$ are quantified. 
\rev{The probabilistic model used in the proposed approach is similar to the one in equation \ref{eq:generative_model_with_projection}. 
However, we adopt a fully Bayesian approach to training its parameters such that full posterior probability distributions for both the GP hyperparameters $\vec{\theta}$ and the projection matrix $\mat{W}$ are obtained. 
As a result, the predictive uncertainty obtained when querying the surrogate model at unobserved input locations accounts for all uncertain model parameters.}

\rev{In view of existing methods~\cite{Tripathy2016, Seshadri2019DimensionFunctions, Rajaram2020Non-IntrusiveSubspace}, the choice of the relevant manifold for the projection matrix $\mat{W}$ remains. 
The Grassmann manifold may appear as the natural choice, since we are seeking a low-dimensional subspace to substitute to the original input space.
However, the \gls{GPR} model with an \gls{ARD} kernel that was chosen to model the link function $g$ grants different length scales to the different input directions of the \gls{GP}.
As such, the particular choice of basis for a given subspace matters when using the \gls{ARD} kernel, not just the subspace.
In other words, the directional information that is retained for elements of the Stiefel manifold but lost for those in the Grassmann manifold is required.
For this reason, we set $\mat{W} \in \mathrm{St}(m, d)$.}

\rev{As opposed to optimization in manifolds, Bayesian inference in manifolds has been developed more recently~\cite{Girolami2009RiemannianCarlo, Betancourt2013GeneralizingManifolds}. Numerical implementations of those algorithms are not yet mature, thus restricting practical Bayesian inference to parameters defined in Euclidean spaces \cite{Bingham2019Pyro:Programming, Ma2013AReductionb}.
In order to accommodate these limitations, reparametrization techniques, that map a set of real-valued parameters to a point on a manifold, have been used~\cite{Nirwan2019RotationPCA}.
In this manner, tools operating with real-valued parameters may be leveraged to perform Bayesian inference in manifolds.
Because we chose to work in the Stiefel manifold, the reparametrization mapping we use associates a vector $\vec{\theta_p}$ of $k$ real parameters to a $d \times m$ orthonormal matrix $\mat{W}$:
\begin{equation}
    \begin{split}
        \mathcal{P}: \mathbb{R}^k &\longrightarrow \mathrm{St}(m, d) \\
        \vec{\theta_p} &\longmapsto \mat{W} = \mathcal{P}(\vec{\theta_p})
    \end{split}
\end{equation}}

The choice of $\mathcal{P}$ is driven by the need to equip $\mat{W}$ with a meaningful prior distribution while recalling that the matching distribution on the parameters $\vec{\theta_p}$ is the one that \rev{must be specified when implementing the probabilistic model.}
\rev{In other words,} along with $\mathcal{P}$, the distribution of $\vec{\theta_p}$ that results in the desired distribution for $\mat{W}$ is needed.
The prior distribution placed on $\mat{W}$ must translate the prior belief that any set of orthonormal directions are \emph{a priori} equally probable candidates, i.e. $p(\mat{W})$ should be a uniform distribution on $\mathrm{St}(m, d)$.
\begin{algorithm}[htbp]
\DontPrintSemicolon
% \SetAlgoLined
\SetKwInOut{Input}{input}\SetKwInOut{Output}{output}
\caption{$\mathcal{H}$: orthonormal matrix parametrization through Householder transformations}\label{alg:householder}
\Input{parameters $\vec{\theta_p} \in \mathbb{R}^k$}
\Output{projection matrix $\mat{W} \in \mathrm{St}(m, d)$}
\BlankLine
$\mat{Q} \leftarrow \mat{I} \in \mathbb{R}^{d \times d}$\;
$l \leftarrow 0$\;
\For{$i\leftarrow 1$ \KwTo $m$}{
$k \leftarrow l$\;
$l \leftarrow k + d - i$\;
$\vec{v} \leftarrow (\theta_{p, k}, \dots, \theta_{p, l})^T$\;
$\vec{u} \leftarrow \dfrac{\vec{v} + \mathrm{sgn}(v_1) || \vec{v} || \vec{e_1}}{ || \vec{v} + \mathrm{sgn}(v_1) || \vec{v} || \vec{e_1} || }$\;
$\mat{\hat{H}} \leftarrow -\mathrm{sgn}(v_1)(\mat{I} - 2 \vec{u} \vec{u}^T)$\;
$\mat{H} \leftarrow \begin{pmatrix}
            \mat{I} & 0 \\
            0 & \mat{\hat{H}}
        \end{pmatrix} $\;
$\mat{Q} \leftarrow \mat{H} \mat{Q}$\;
}
$\mat{W} \leftarrow (\vec{Q_1}, \dots, \vec{Q_m}) \in \mathbb{R}^{d \times m}$\;
\end{algorithm}

\rev{The Stiefel manifold can be endowed with a uniform measure that is a Haar measure, i.e. it remains unchanged by the application of orthogonal transformations: $p(\mat{W}) = p(\mat{QW})\ \forall \mat{Q} \in O(d)$ where $O(d)$ is the orthogonal group~\cite{Nirwan2019RotationPCA}.
As shown in \cite{Nirwan2019RotationPCA}, if the projection parameters $\vec{\theta_p}$ are i.i.d. Gaussian random variables and the Householder parametrization $\mathcal{H}$ detailed in algorithm \ref{alg:householder} is used for the reparametrization mapping $\mathcal{P}$ such that $\mat{W} = \mathcal{H}(\vec{\theta_p})$, then $\mat{W}$ is a random orthogonal matrix with distribution given by the Haar measure in $\mathrm{St}(m, d)$, which is the desired prior distribution for $\mat{W}$.}

\rev{Compared to other reparametrization methods, the Householder transformation has the advantage of not requiring a computationally burdensome change of measure \cite{Nirwan2019RotationPCA}.
The number $k$ of real-valued parameters $\vec{\theta_p}$ is $k = md - m(m-1)/2$, which is larger than the actual dimension of the Stiefel manifold: $\mathrm{dim}(\mathrm{St}(m, d))) = md - m(m+1)/2$. 
Doing without a change of measure comes at the cost of $m$ additional parameters. 
This is not a significant penalty in practice since dimension reduction methods seek a low-dimension space such that $m \ll d$.}

\rev{Prior distributions for the remaining model parameters must also be specified: a log-normal distribution is used as prior for the \gls{GP} hyperparameters. 
This results the proposed generative model detailed in algorithm \ref{alg:fully_bayesian_model}.} 
\begin{algorithm}[htbp]
\DontPrintSemicolon
\caption{Proposed Fully Bayesian Model} \label{alg:fully_bayesian_model}
$\theta_{p,i} \sim \mathcal{N}(0,1) \quad \forall i = 1, \dots, k \label{eq:prior_proj_params}$\; 
$\mat{W} = \mathcal{H}(\vec{\theta_p}) \quad \mathrm{(algorithm\ \ref{alg:householder})}$\;
$\mat{Z} = \mat{X} \mat{W}$\;
$\log{\theta_j} \sim \mathcal{N}(0,1) \quad \forall j = 1, \dots, m\rev{+2}$\;
$\vec{y} \sim \mathcal{N}(0, K(\mat{Z}, \mat{Z}; \vec{\theta}) + \sigma_n^2 \mat{I}) \label{eq:low_dim_gp}$\;
\end{algorithm}

\rev{The training of the proposed model consists in inferring} the joint posterior distribution $p(\vec{\theta}, \vec{\theta_p} | \mat{X}, \vec{y})$ of the model parameters by conditioning the generative model shown in algorithm \ref{alg:fully_bayesian_model} with respect to the training data $\mathcal{D} = (\mat{X}, \vec{y})$ and performing probabilistic inference using \gls{MCMC}.

\rev{Once posterior distributions are obtained, the model can be used to make predictions.}
In contrast to traditional \gls{GPR} (equation \eqref{eq:bayesian_gp_predictive}), fully Bayesian predictions in the proposed approach require marginalizing over both the GP hyperparameters and the projection parameters:
\begin{equation} \label{eq:bayesian_as_gp_predictive}
    p(\vec{y_*} | \mat{X_*}, \mat{X}, \vec{y}) = \iint p(\vec{y_*} | \mat{X_*}, \mat{X}, \vec{y}, \vec{\theta}, \vec{\theta_p}) p(\vec{\theta}, \vec{\theta_p} | \mat{X}, \vec{y}) \, d{\vec{\theta}} \, d{\vec{\theta_p}}
\end{equation}
where equation \eqref{eq:posterior_predictive} is used to compute the term $p(\vec{y_*} | \mat{X_*}, \mat{X}, \vec{y}, \vec{\theta}, \vec{\theta_p})$ after projecting the input design matrix $\mat{X}$ onto the \rev{feature space}.

In addition to the prediction of the response $\vec{y_*}$ at unobserved locations $\mat{X_*}$, the posterior distribution $p(\mat{W} \vert \mat{X}, \vec{y})$ of the projection matrix $\mat{W}$ can be readily obtained by application of the Householder parametrization detailed in algorithm \ref{alg:householder} on the \gls{MCMC} samples of the marginal distribution $p(\vec{\theta_p} | \mat{X}, \vec{y})$.
Instead of a point-based estimation, the proposed method therefore grants access to a \rev{full posterior probability distribution of the projection matrix $\mat{W}$}. 

We will refer from now on refer to the approach presented in this section as \gls{BAS}.

\section{Results}
\label{sec:results}
\rev{In this section, we present and discuss results obtained through the application of the proposed approach to a variety of examples. It is organized as follows: in section \ref{sec:experimental_setup}, we start by introducing the datasets and evaluation metrics we will rely on throughout the presentation. Then, section \ref{sec:in_depth_run_through} focuses on the application of the method on a single dataset, and detailed metrics pertaining to the training and validation processes are shown. Finally, in section \ref{sec:comparative_study}, we compare the predictive performance of the proposed approach to two other state-of-the-art benchmark methods.}
\subsection{Experimental Setup}
\label{sec:experimental_setup}
\subsubsection{Benchmark Datasets}
\begin{table}[ht]
    \centering
    \begin{tabular}{cl|cccc}
\toprule
 & & Sample Size & \makecell{Input Space\\ Dimension} & \makecell{Active Subspace\\ Dimension} & Reference \\
\midrule
\multirow{8}{*}{\makecell{Analytical Quadratic\\ Functions (QF)}} 
& QF 10/1         & 1000          & 10                & 1 & -- \\
& QF 10/2         & 1000          & 10                & 2 & -- \\
& QF 25/1         & 1000          & 25                & 1 & -- \\
& QF 25/2         & 1000          & 25                & 2 & -- \\
& QF 50/1         & 1000          & 50                & 1 & -- \\
& QF 50/2         & 1000          & 50                & 2 & -- \\
& QF 100/1        & 1000          & 100               & 1 & -- \\
& QF 100/2        & 1000          & 100               & 2 & -- \\
\midrule
\multirow{4}{*}{\makecell{Science and\\ Engineering}} 
& NACA0012 (lift) & 1756          & 18                & 1 & \cite{Constantine2015ActiveStudies} \\
& HIV at $t=3400$ & 1000          & 27                & 1 & \cite{Loudon2017MathematicalHIV} \\
& ONERA M6 (lift) & 297           & 50                & 1 & \cite{Lukaczyk2014ActiveOptimization} \\
& Elliptic PDE    & 1000          & 100               & 1 & \cite{Constantine2016AcceleratingSubspaces} \\
\bottomrule
\end{tabular}
\caption{Summary of benchmark datasets}
\label{tab:datasets_summary}
\end{table}

The comparison of the proposed method's performance with benchmark method is drawn based on datasets generated using analytical functions and on datasets originating from science and engineering. Their features are summarized in table \ref{tab:datasets_summary}.

Analytical functions are chosen to be quadratic functions featuring a \rev{dependence on a low-dimensional input subspace} by construction. As before, $d$ and $m$ are respectively the dimensions of the input space and \rev{feature space}. The quadratic functions $f$ are defined as:
\begin{align}
    \vec{z} &= \mat{W}^T \vec{x} \\
    f(\vec{x}) &= \vec{z}^T \mat{A} \vec{z} + \vec{b} \vec{z} + c + \epsilon
\end{align}
where $\vec{x}$ and $\vec{z}$ are respectively the input vector and the vector of projected coordinates in the \gls{FS}. 
$\mat{W}$ is the $d \times m$ projection matrix onto the \gls{FS}. $\mat{A} \in \mathbb{R}^{m \times m}$, $\vec{b} \in \mathbb{R}^{m}$, and $c \in \mathbb{R}$ are parameters of the quadratic mappings. 
Given $d$ and $m$, their coefficients are sampled from standard normal distributions to obtain randomly generated mappings. 
Since the domains of the quadratic mappings are restricted to their respective \gls{FS} by construction, an additive centered Gaussian noise $\epsilon$ of standard deviation $5 \times 10^{-2}$ is added to simulate the variation of the response due to variations of the inputs in the inactive subspace. 
In the context of the present study, eight quadratic functions were generated, that combine an input space dimension $d$ of 10, 25, 50, or 100 with \rev{an \gls{FS} dimension of 2 or 5}.

Four datasets originating from scientific and engineering applications are also used to assess the proposed method. These datasets were previously used in studies \rev{related to the \gls{AS} method} and include gradient evaluations in addition to input-output pairs. Gradient evaluations are used to estimate the actual \gls{AS} using the original \gls{AS} method, which is needed to assess the ability of the proposed method to uncover the \gls{AS}. Those datasets cover a wide range of input space dimensions: the NACA 0012~\cite{Constantine2015ActiveStudies}, HIV~\cite{Loudon2017MathematicalHIV}, ONERA-M6~\cite{Lukaczyk2014ActiveOptimization}, and Elliptic PDE~\cite{Constantine2016AcceleratingSubspaces} datasets respectively have 18, 27, 50, and 100 input dimensions. Based on previous studies, all of these datasets are known to have a one-dimensional \gls{AS}.

\subsubsection{Benchmark Metrics}
\rev{The comparative study focuses on the four following aspects: a) deterministic predictive capability, b) probabilistic predictive capability, c) computational cost, and d) similarity of the uncovered subspace with the \gls{AS}. To each aspect corresponds a numerical metric enabling the quantitative comparison of the proposed method with both benchmark methods: a) \glsentrylong{r2}, b) \glsentrylong{MLPPD}, and c) \glsentrylong{TD}, and d) subspace angles. These metrics are briefly defined in the following paragraphs.}
\paragraph{Coefficient of Determination}
\rev{The deterministic predictive capability refers to the quality of point predictions made by the model and is quantified using the \gls{r2} metric defined below.
\begin{definition}[\Glsxtrfull{r2}]
Let $n$ test points $\{\vec{x^*_1}, ..., \vec{x^*_n}\}$ distributed according to the distribution $p(\vec{x})$, $f$ the underlying function of interest, and $\hat{f}$ its approximation, the \glsxtrfull{r2} is:
\begin{equation} \label{eq:r_squared}
    1 - \frac{\frac{1}{n} \sum_{i=1}^{n}{ \left( f(\vec{x^*_i}) - \hat{f}(\vec{x^*_i}) \right)^2}}{\frac{1}{n} \sum_{i=1}^{n}{f(\vec{x^*_i})^2}}
\end{equation}
\end{definition}
For fully Bayesian methods, the point estimate $\hat{f}(\vec{x^*})$ at a new location $\vec{x^*}$ is chosen as the median of the posterior predictive distribution $p(y_* | \vec{x_*}, \mat{X}, \vec{y})$ introduced in equation \eqref{eq:bayesian_as_gp_predictive}. 
The median is approximated as follows.
Samples from the joint distribution $p(y_*, \vec{\theta}, \vec{\theta_p} | \vec{x_*}, \mat{X}, \vec{y})$ are drawn using the fact that:
\begin{equation}
    p(y_*, \vec{\theta}, \vec{\theta_p} | \vec{x_*}, \mat{X}, \vec{y}) = p(y_* | \mat{X_*}, \mat{X}, \vec{y}, \vec{\theta}, \vec{\theta_p}) p(\vec{\theta}, \vec{\theta_p} | \mat{X}, \vec{y})
\end{equation}
Samples from $p(\vec{\theta}, \vec{\theta_p} | \mat{X}, \vec{y})$ are byproducts of the MCMC process and $p(y_* | \mat{X_*}, \mat{X}, \vec{y}, \vec{\theta}, \vec{\theta_p})$ is a known Gaussian distribution.
For every draw from $p(\vec{\theta}, \vec{\theta_p} | \mat{X}, \vec{y})$, multiple draws from $p(y_* | \mat{X_*}, \mat{X}, \vec{y}, \vec{\theta}, \vec{\theta_p})$ are performed.
This results into a collection of draws from the joint distribution $p(y_*, \vec{\theta}, \vec{\theta_p} | \vec{x_*}, \mat{X}, \vec{y})$. 
Information regarding $\vec{\theta}$ and $\vec{\theta_p}$ is dropped from these samples, effectively marginalizing out those parameters, and yielding draws from the desired posterior predictive distribution $p(y_* | \vec{x_*}, \mat{X}, \vec{y})$.
The median of $p(y_* | \vec{x_*}, \mat{X}, \vec{y})$ is approximated by the sample median of these draws.}
\paragraph{Mean Log Pointwise Predictive Density}
\rev{The probabilistic predictive capability refers to the quality of the probability distributions outputted by the model. Because the computer models under consideration are deterministic mappings, the likelihood of observing the actual value output under the posterior predictive distribution can be used as a metric. The log pointwise predictive density metric \cite{bayesiandataanalysis} can be used for this purpose. To account for different numbers of validation samples across cases, the metric is normalized by taking the mean over the observations used for its computation. We refer to the resulting metric as \glsentryfull{MLPPD}, it is defined below.}
\rev{\begin{definition}[\Glsxtrfull{MLPPD}]
Let $n$ test points $\{\vec{x^*_1}, ..., \vec{x^*_n}\}$ be distributed according to the distribution $p(\vec{x})$, the \gls{MLPPD} is:
\begin{equation}
    \frac{1}{n} \sum_{i=1}^{n}{\log{p(y_i^*|\vec{x_i^*}, \mat{X}, \vec{y})}}
\end{equation}
\end{definition}}
\rev{The expression for the posterior predictive distribution was recalled in equation \eqref{eq:posterior_predictive}. With $\mu_i^*$ and ${\sigma_i^*}^2$ respectively computed using equations \eqref{eq:mu_star} and \eqref{eq:Sigma_star}, we obtain the following expression:
\begin{equation}
    \log{p(y_i^*|\vec{x_i^*}, \mat{X}, \vec{y})} = 
    -\frac{1}{2} \frac{(y_i^* - \mu_i^*) ^ 2}{{\sigma_i^*}^2}
    - \log{\sigma_i^*} 
    - \frac{\gamma}{2} \log{2 \pi}
\end{equation}
where $\gamma = m$ for the \gls{MOAS} and \gls{BAS} methods and $\gamma = d$ for the \gls{BAS} method.}

\rev{For fully Bayesian methods, marginalization with respect to the hyperparameters is necessary:
\begin{equation} \label{eq:bayesian_posterior_validation_lok_lik}
    \log{p(y_i^*|\vec{x_i^*}, \mat{X}, \vec{y})} 
    = \log{\left(
        \iint{
            p(y_i^*|\vec{x_i^*}, \mat{X}, \vec{y}, \vec{\theta}, \vec{\theta_p})
            p(\vec{\theta}, \vec{\theta_p} | \mat{X}, \vec{y}) 
            \, d{\vec{\theta_p}}
            \, d{\vec{\theta}}
        }
    \right) 
    }
\end{equation}
Computation of the double integral in equation \eqref{eq:bayesian_posterior_validation_lok_lik} relies on a Monte-Carlo approximation that uses the samples from the joint posterior distribution $p(\vec{\theta}, \vec{\theta_p} | \mat{X}, \vec{y})$ obtained through \gls{MCMC}.}
\paragraph{Training Duration}
\rev{\begin{definition}[\Gls{TD}]
\Glsxtrlong{TD} is measured as the wall-clock time elapsed between the start and the end of the model training.
\end{definition}}
\paragraph{Mean First Subspace Angle}
\rev{\gls{r2} and \gls{MLPPD} allow to assess the performance of the proposed approach with respect to its end application, namely creating a surrogate model that is an accurate image of the original mapping of interest and that provides accurate quantification of epistemic uncertainty due to limited data. 
However, they do not provide any insight regarding the inner workings of the proposed approach.
In particular, the feature space on which original inputs are projected to serve as low-dimensional inputs to a \gls{GP} is hidden by those metrics.
A direct assessment of the subspace is not straightforward, as methods seeking a ridge approximation of the form $f(\vec{x}) = g(\mat{W}^T \vec{x})$ may yield different projection matrices $\mat{W}$ and subspaces \cite{Constantine2017AApproximation}.
Despite those facts, the comparison of the uncovered feature space with the subspace yielded by benchmark methods, and in particular the \gls{AS} method, has proven to be insightful. 
In many instances, we show that the proposed approach eventually recovers the \gls{AS} once a sufficient training number of training data is used. 
In those instances, the comparison of the uncovered feature space with the \gls{AS} enables to link the poor predictive performance with smaller training sets to the inability of the method to uncover an adequate subspace.}

\rev{Directly comparing projection matrices $\mat{W}$ does not allow to draw conclusions regarding the subspaces they span, as infinitely many orthogonal bases may be obtained through rotations within the subspace. 
Instead, principal angles between subspaces, or simply subspace angles, provide a similarity measure between subspaces that does not depend on the particular choice of bases for these subspaces.
The definition of subspace angles from \cite{Knyazev2012PrincipalTangents} is reproduced in definition~\ref{def:subspace_angles}.
\begin{definition}[Definition 2.1. in \cite{Knyazev2012PrincipalTangents}]
\label{def:subspace_angles}
Let $\mathcal{X} \subset \mathbb{C}^n$ and $\mathcal{Y} \subset \mathbb{C}^n$ be subspaces with $\mathrm{dim}(\mathcal{X}) = p$ and $\mathrm{dim}(\mathcal{Y}) = q$. 
Let $m = \mathrm{min}(p, q)$. 
The principal angles
\begin{equation*}
\Theta(\mathcal{X}, \mathcal{Y}) = [\theta_1, \dots, \theta_m], \mathrm{where\ } \theta_k \in [0, \pi/2], k = 1, \dots, m,
\end{equation*}
between $\mathcal{X}$ and $\mathcal{Y}$ are recursively defined by
\begin{equation*}
s_k = \mathrm{cos}(\theta_k) = \max_{x \in \mathcal{X}} \max_{y \in \mathcal{Y}}  \left| x^H y \right| = \left| x_k^H y_k \right|,
\end{equation*}
subject to
\begin{equation*}
\|x\| = \|y\| = 1, x^H x_i = 0, y^H y_i = 0, i = 1, \dots, k-1.
\end{equation*}
The vectors $\{x_1, \dots, x_m\}$ and $\{y_1, \dots, y_m\}$ are called the principal vectors.
\end{definition}
By this definition, the first subspace angle is greater than the following angles.
It is therefore an upper bound for all subspace angles. 
For this reason and because using a single numerical value eases comparison between methods, the first subspace angle will be used as metric. }

\rev{The reference \gls{AS} is computed using the original \gls{AS} method \cite{Constantine2015ActiveStudies} presented in section \ref{sec:active_subspace}. This is possible because all test datasets include gradient evaluations. For each dataset, all available gradient samples are used to estimate the reference \gls{AS} in an effort to maximize the quality of the \gls{MC} estimator $\sum_{i=0}^{n}{\nabla f(\vec{x_i}) {\nabla f(\vec{x_i})}^T}$.}

\rev{The probability distribution for the projection matrix $\mat{W}$ that we obtain using Bayesian inference leads to a distribution on the subspace angles between the recovered feature space and the \gls{AS} computed using gradients. To ease the presentation of the results, we use its mean as metric instead of the full distribution and we refer to it as the \gls{MFSA}.}
\rev{\subsection{In-Depth Walk-Through}}
\label{sec:in_depth_run_through}
\rev{This section presents a detailed walk-through of the application of the proposed method on an engineering use case. 
After a brief presentation of the problem, we examine the training phase of the model, and then present an assessment of the surrogate's predictive performance.
We focus on a single of the benchmark datasets presented in the previous section, namely the ONERA-M6 dataset from~\cite{Lukaczyk2014ActiveOptimization}. 
In this problem, the authors sought to predict the impact of shape deformations encoded by 50 \gls{FFD} control points on the lift produced by an ONERA-M6 wing.}

\rev{The quality of the model parameters' inference and the final model's predictive performance are affected by multiple factors, including the dimension of the \gls{FS} and the number of observations on which the model is conditioned. 
These parameters have been varied in our experiments to understand their impact. 
The dimension of the \gls{FS} was varied from 1 to 5 and the the number of training samples has been varied from 1 to 5 times the number of input dimensions.
For conciseness, we will only present results for notional "low" and "high" data regimes, respectively corresponding to 100 and 250 training samples for the current 50-dimensional problem.}

\begin{figure}%
    \centering
    \subfigure[1D feature space / 100 training samples]{\includegraphics{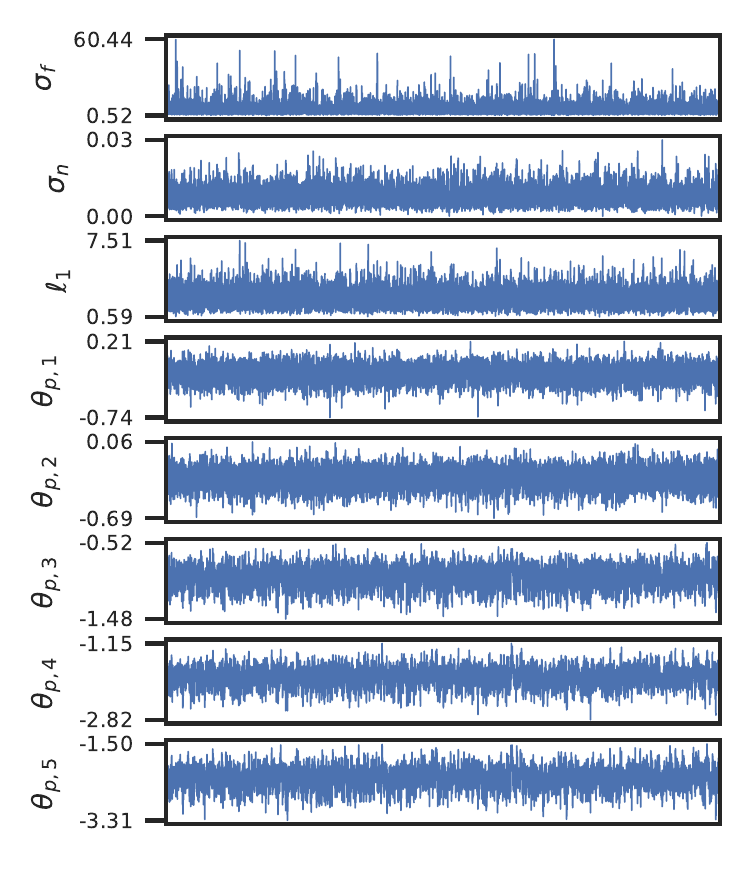}}\qquad
    \subfigure[1D feature space / 250 training samples]{\includegraphics{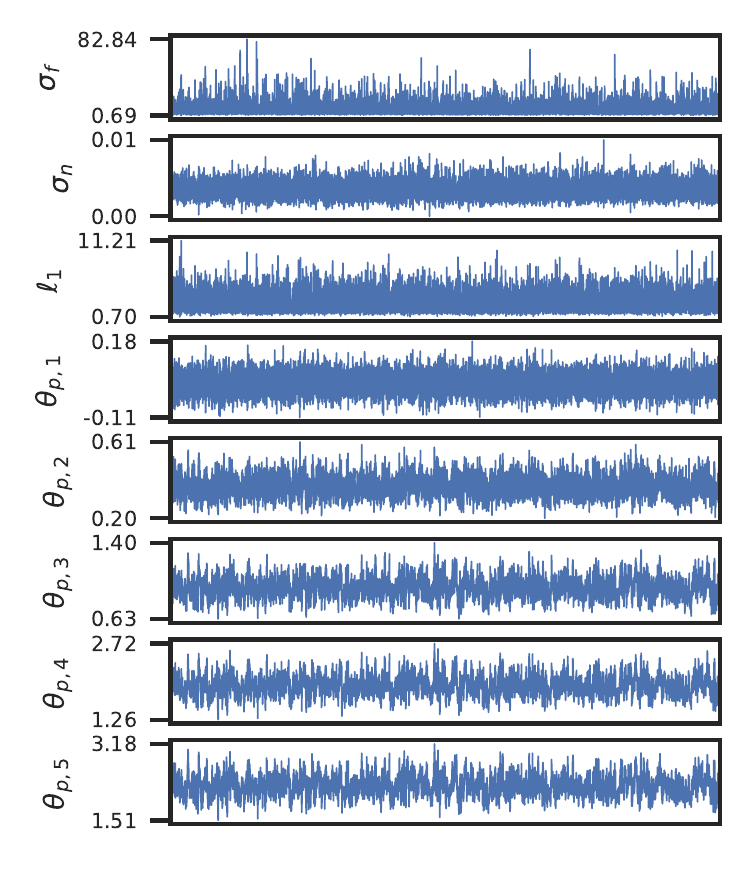}}\\
    \subfigure[5D feature space / 100 training samples]{\includegraphics{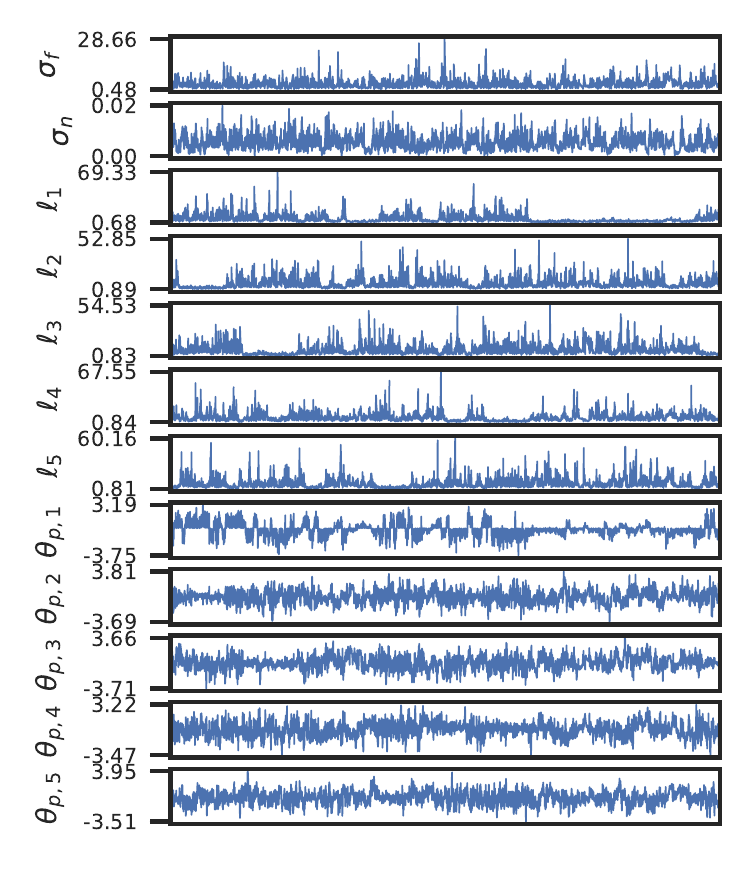}}\qquad
    \subfigure[5D feature space / 250 training samples]{\includegraphics{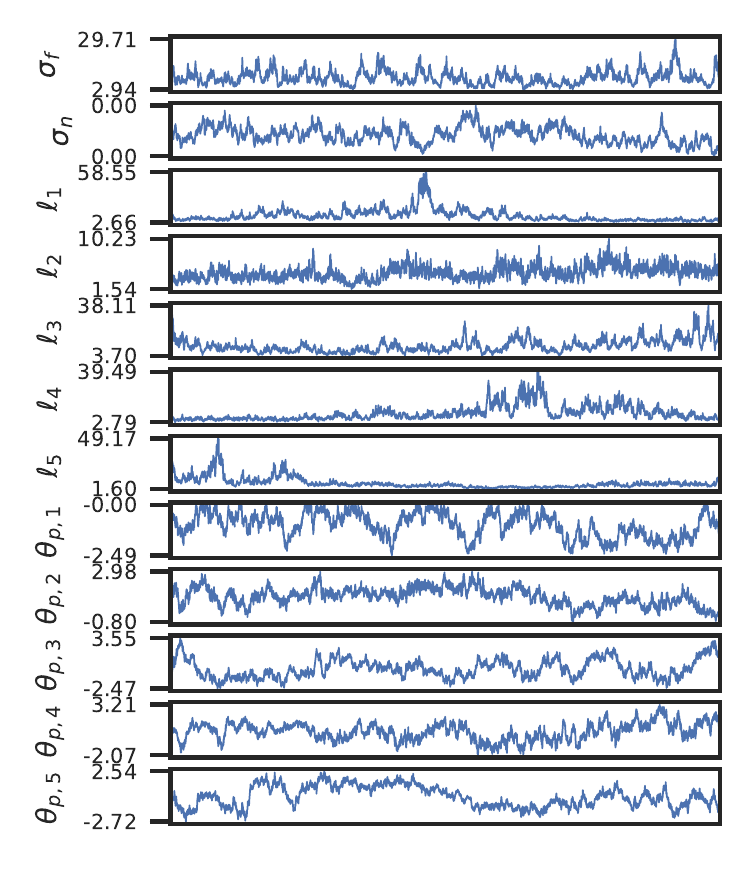}}%
    \caption{Markov-Chain Monte-Carlo chains of the model parameters for four different combinations of feature space dimension and number of training samples. Each chain contains 1,000 draws. Only the first five projection parameters are shown for brevity.}
    \label{fig:mcmc_chains}
\end{figure}

\begin{figure}%
    \centering
    \subfigure[Number of training samples fixed to 100, x-axis is number of feature space dimensions.]{\includegraphics{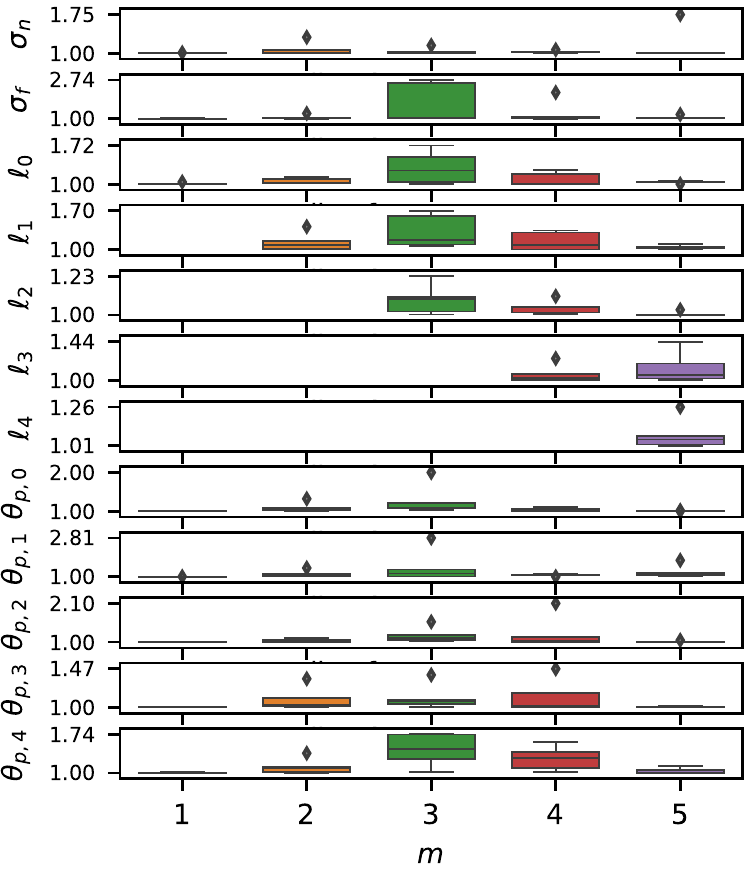}}\qquad
    \subfigure[Number of training samples fixed to 250, x-axis is number of feature space dimensions.]{\includegraphics{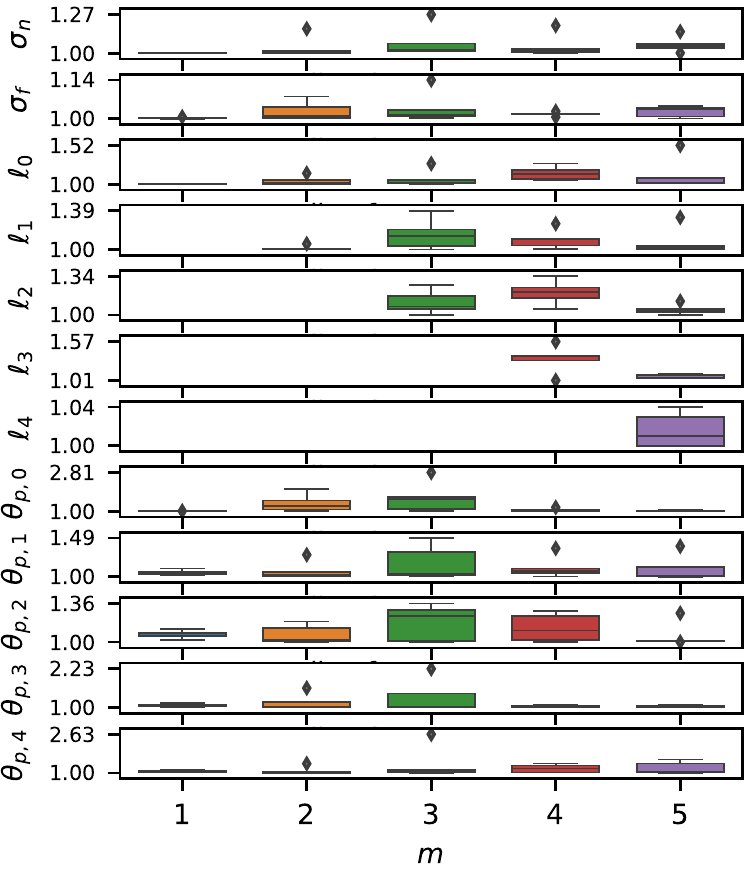}}\\
    \subfigure[Feature space dimension fixed to 1, x-axis is number of training samples.]{\includegraphics{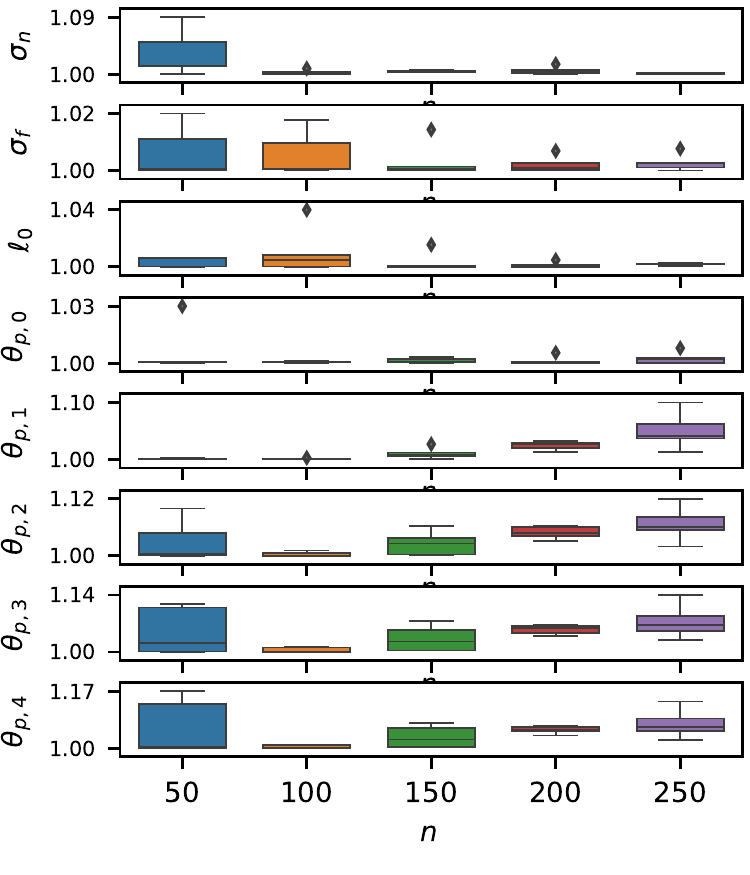}}\qquad
    \subfigure[Feature space dimension fixed to 5, x-axis is number of training samples.]{\includegraphics{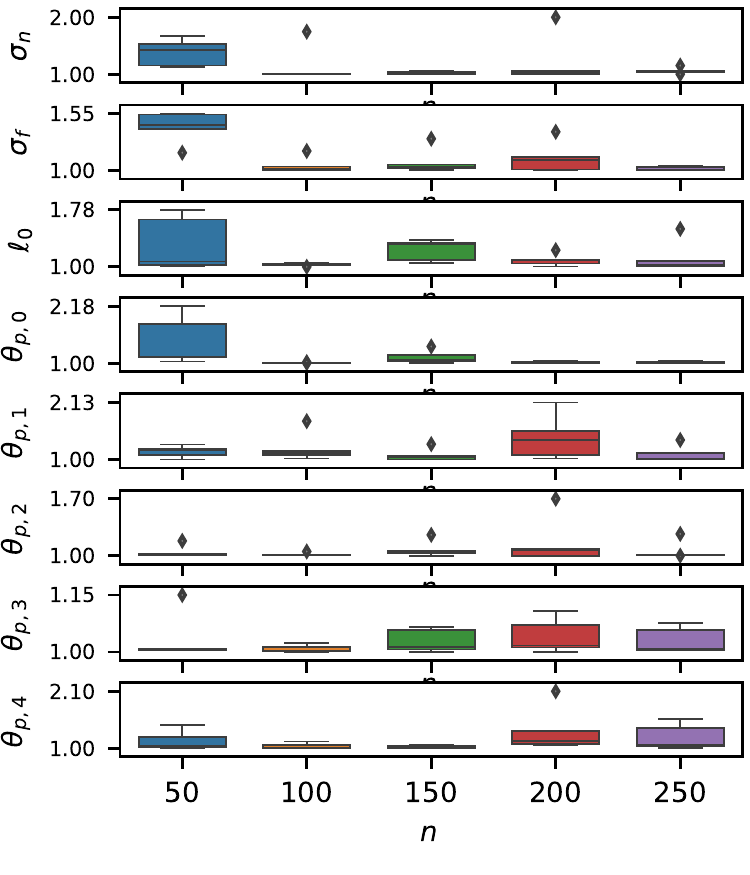}}%
    \caption{Values of the split Gelman-Rubin statistics of the \gls{MCMC} chains as a function of the feature space dimension $m$ for two fixed numbers of training samples (a, b), and as a function of the number $n$ of training samples for two fixed feature space dimensions (c, d). Parameter names are indicated to the left of the plot.}
    \label{fig:split_gelman_rubin}
\end{figure}

\begin{figure}%
    \centering
    \subfigure[1D feature space / 100 training samples]{\includegraphics{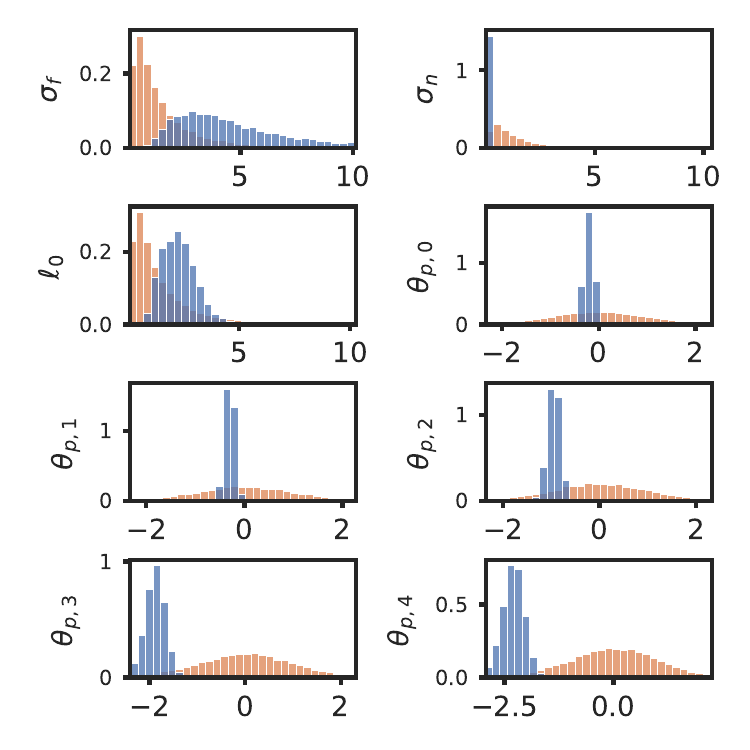}}\qquad
    \subfigure[1D feature space / 250 training samples]{\includegraphics{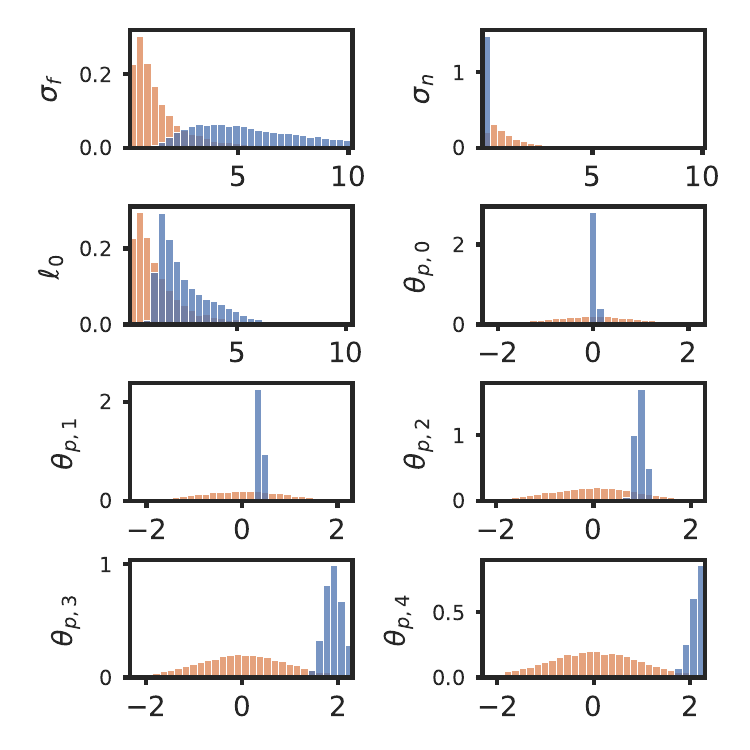}}\\
    \subfigure[5D feature space / 100 training samples]{\includegraphics{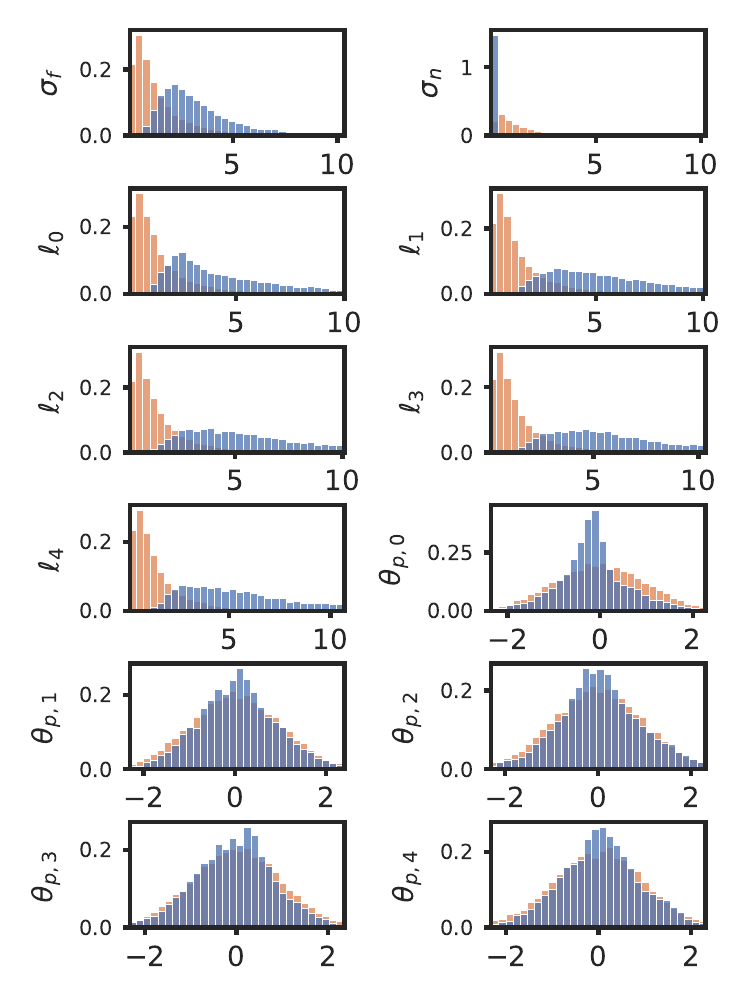}}\qquad
    \subfigure[5D feature space / 250 training samples]{\includegraphics{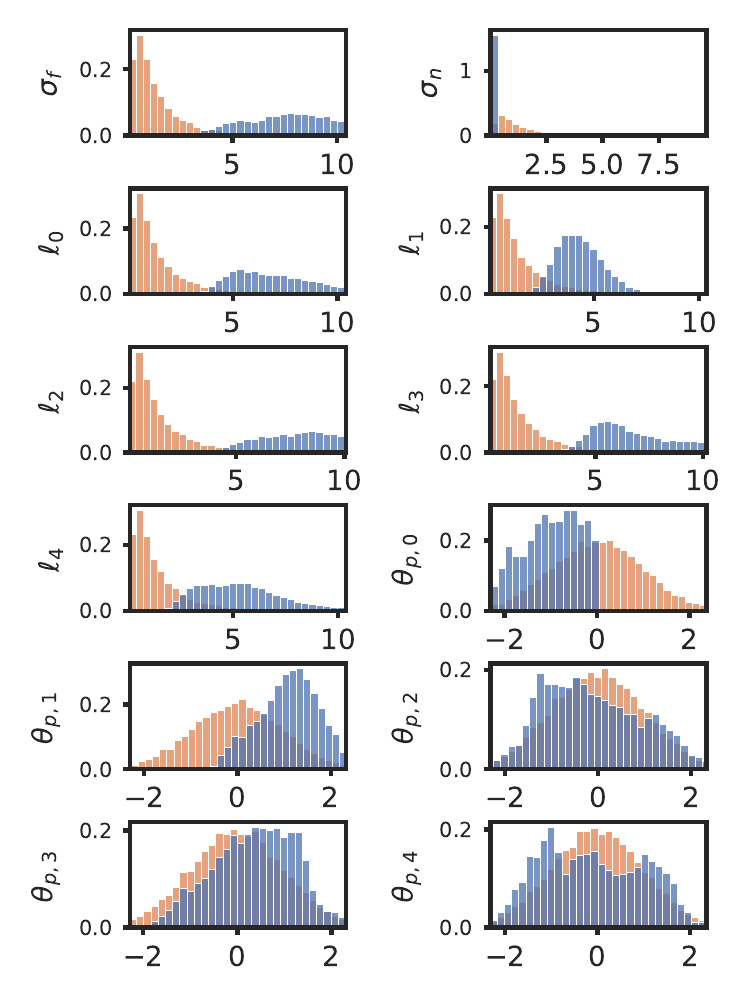}}%
    \caption{Prior (orange) and posterior (blue) distributions. Parameter names are indicated to the left of the plot. Histograms are normalized such that their respective areas equal 1.}
    \label{fig:prior_vs_posterior}
\end{figure}

\begin{figure}%
    \centering
    \subfigure[1D feature space / 100 training samples]{\includegraphics{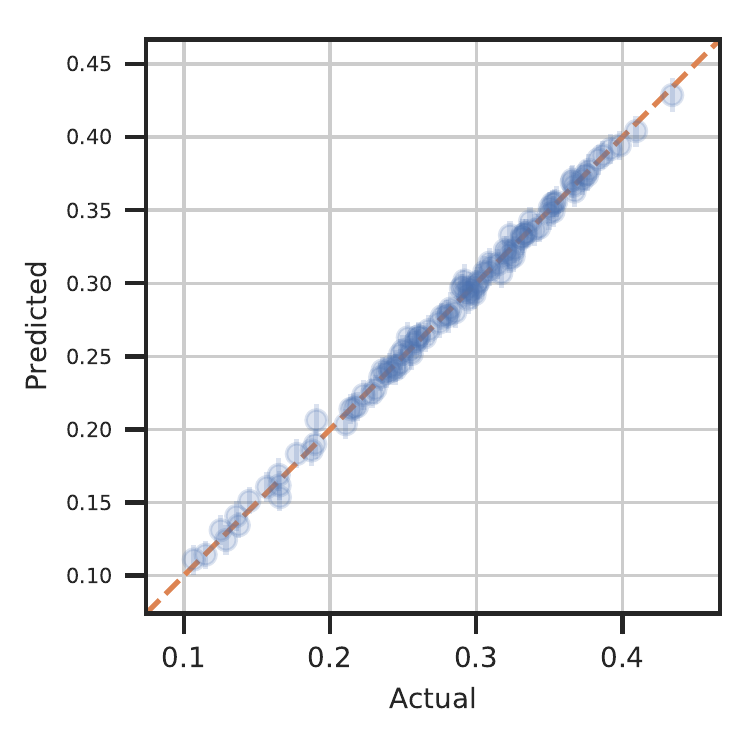}}\qquad
    \subfigure[1D feature space / 250 training samples]{\includegraphics{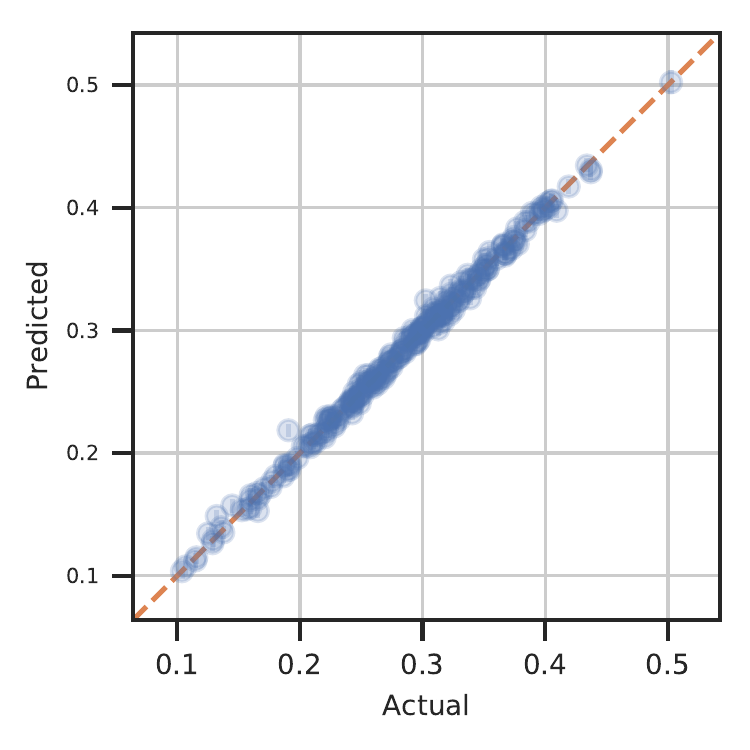}}\\
    \subfigure[5D feature space / 100 training samples]{\includegraphics{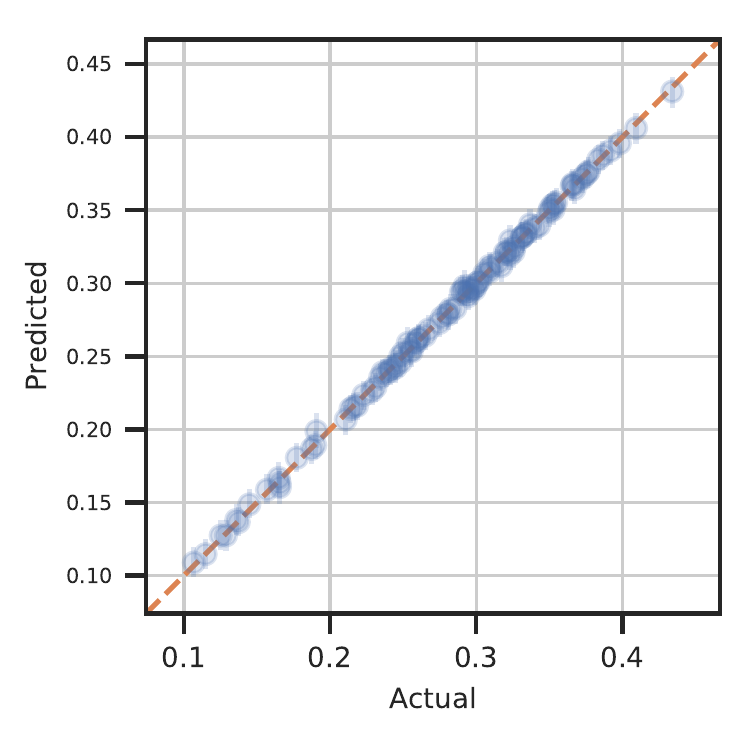}}\qquad
    \subfigure[5D feature space / 250 training samples]{\includegraphics{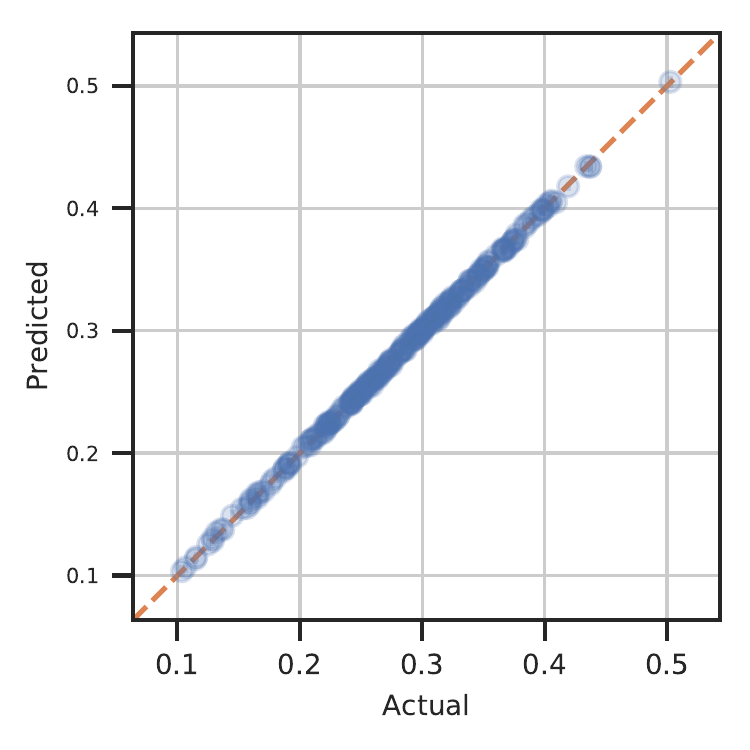}}%
    \caption{Comparison of the training data (Actual) and the model predictions (Predicted) for the ONERA M6 dataset for four different combinations of feature space dimension and number of training samples. Vertical bars indicate the 95\% confidence interval.}
    \label{fig:training_actual_vs_predicted}
\end{figure}

\begin{figure}%
    \centering
    \subfigure[Number of training samples fixed to 100, x-axis is number of feature space dimensions.]{\includegraphics{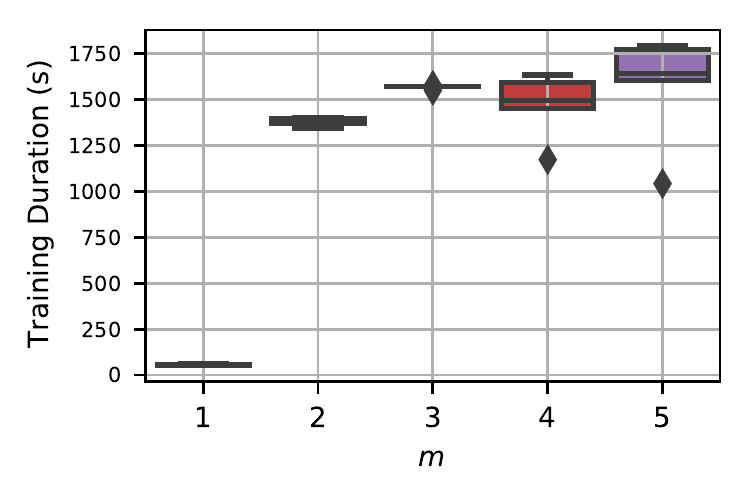}}\qquad
    \subfigure[Number of training samples fixed to 250, x-axis is number of feature space dimensions.]{\includegraphics{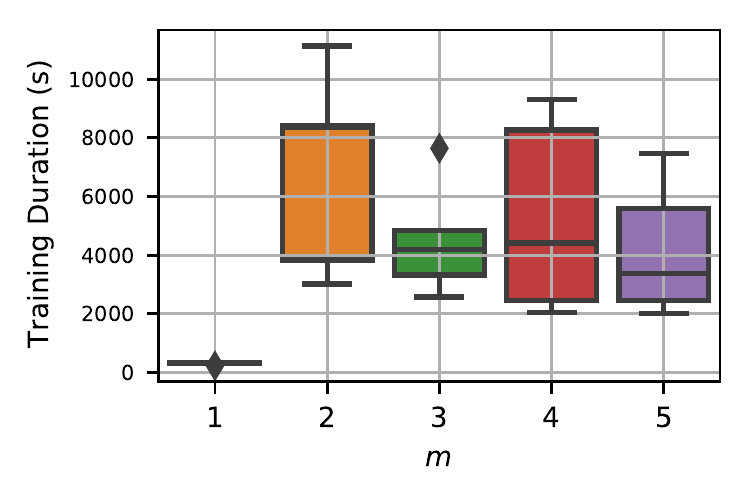}}\\
    \subfigure[Feature space dimension fixed to 1, x-axis is number of training samples.]{\includegraphics{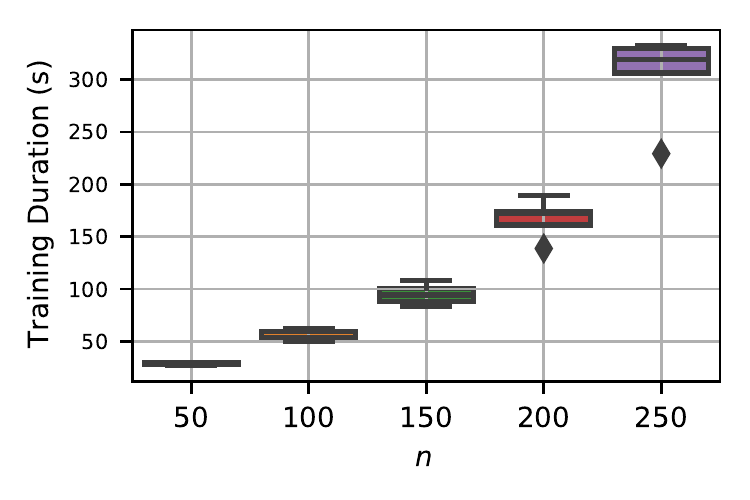}}\qquad
    \subfigure[Feature space dimension fixed to 5, x-axis is number of training samples.]{\includegraphics{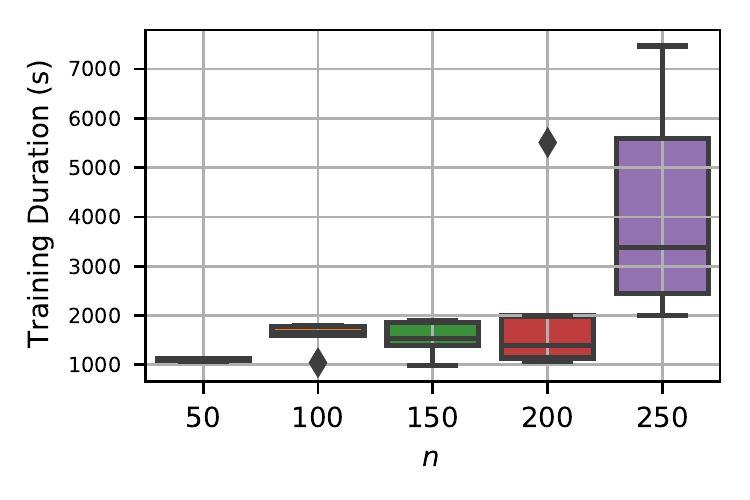}}\\
    \caption{Training duration as a function of the feature space dimension $m$ for two fixed numbers of training samples (a, b), and as a function of the number $n$ of training samples for two fixed feature space dimensions (c, d).}
    \label{fig:training_duration}
\end{figure}

\begin{figure}%
    \centering
    \subfigure[1D feature space / 100 training samples]{\includegraphics{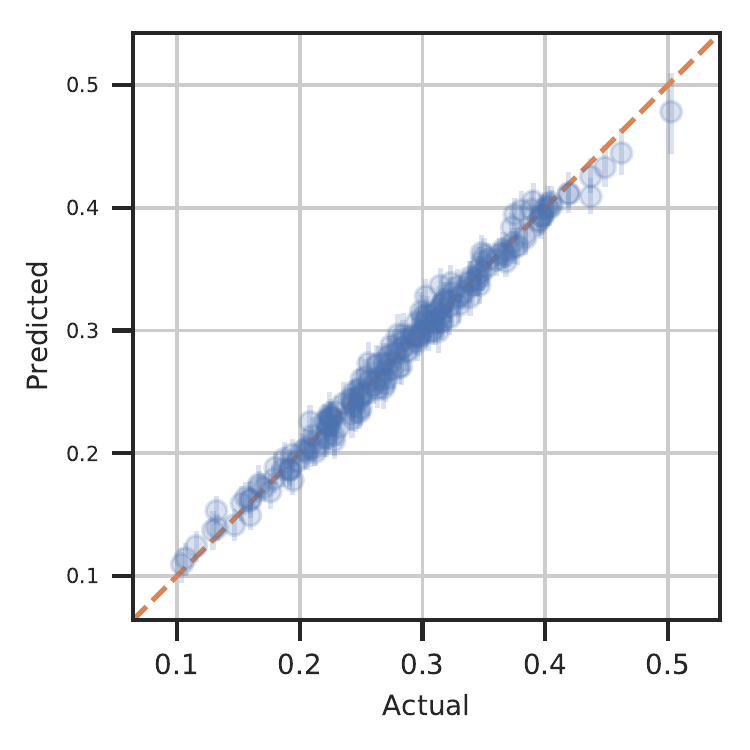}}\qquad
    \subfigure[1D feature space / 250 training samples]{\includegraphics{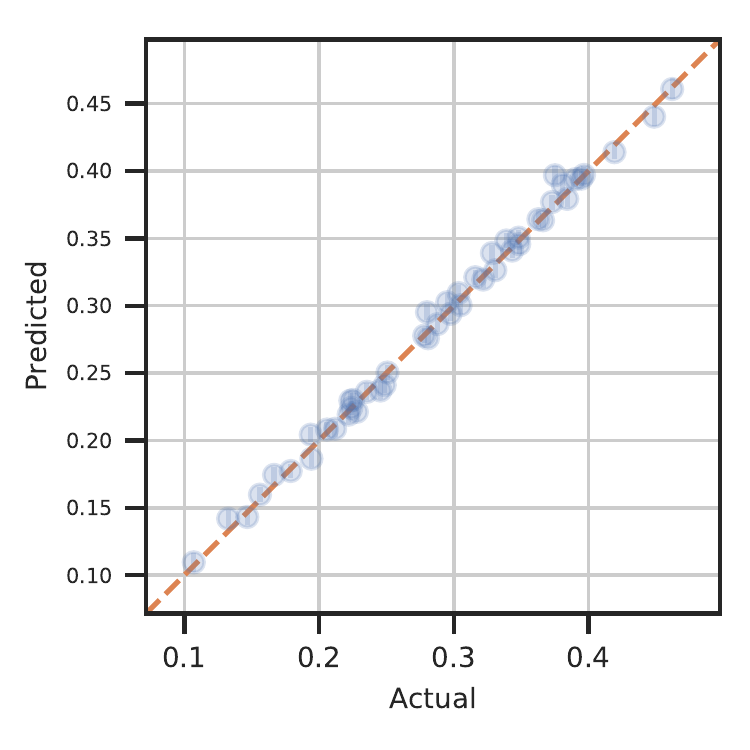}}\\
    \subfigure[5D feature space / 100 training samples]{\includegraphics{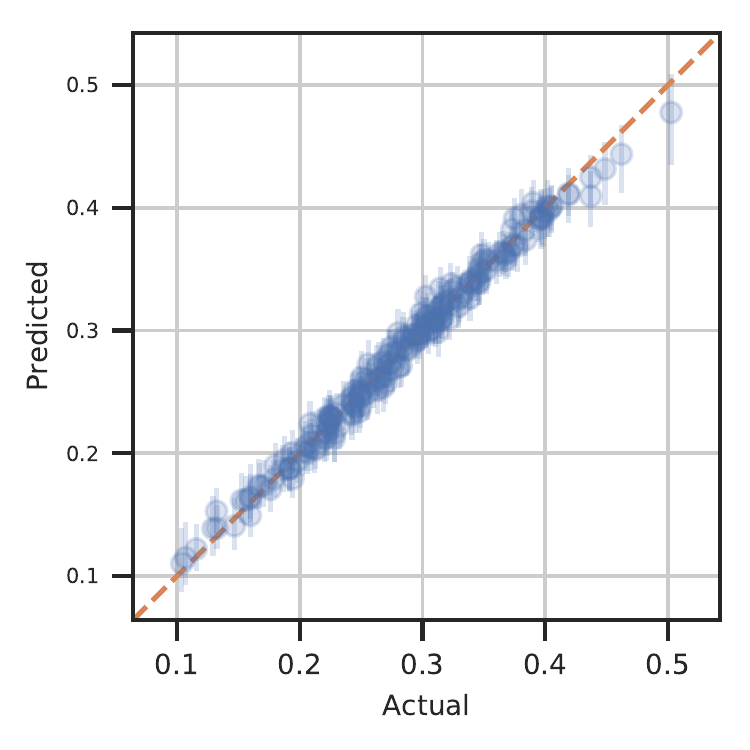}}\qquad
    \subfigure[5D feature space / 250 training samples]{\includegraphics{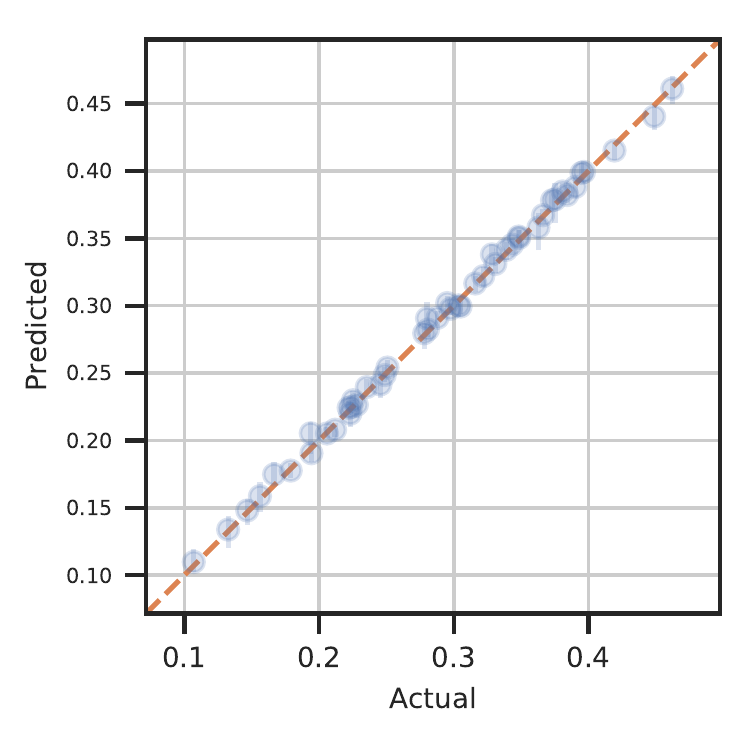}}%
    \caption{Comparison of the actual dataset and the model predictions for validations points, i.e. observations not used during training. Comparisons are shown for four different combinations of feature space dimension and number of training samples. Vertical bars indicate the 95\% confidence interval.}
    \label{fig:validation_actual_vs_predicted}
\end{figure}

\begin{figure}%
    \centering
    \subfigure[Number of training samples fixed to 100, x-axis is number of feature space dimensions.]{\includegraphics{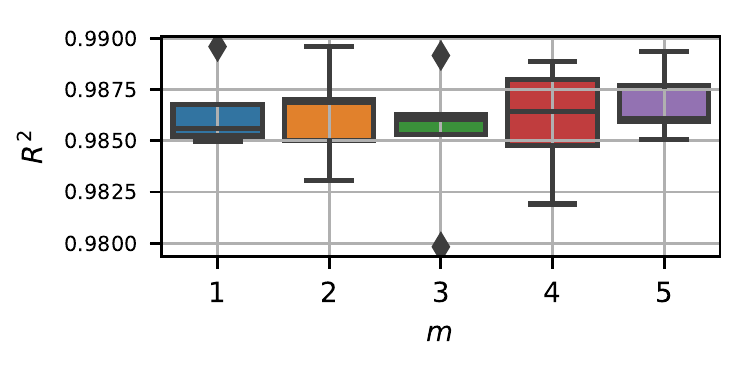}}\label{r_squared_vs_m_100TS}\qquad
    \subfigure[Number of training samples fixed to 250, x-axis is number of feature space dimensions.]{\includegraphics{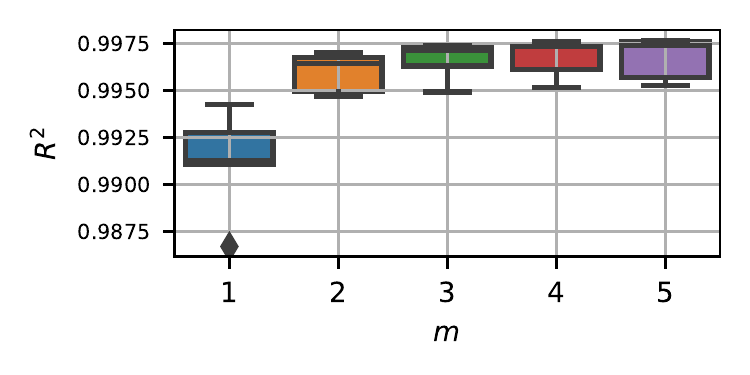}}\\
    \subfigure[Feature space dimension fixed to 1, x-axis is number of training samples.]{\includegraphics{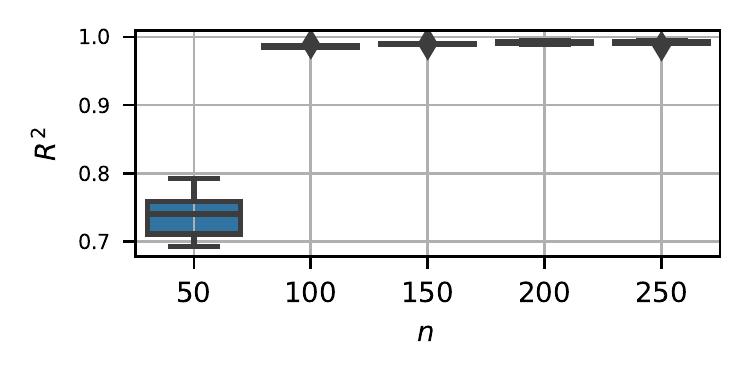}}\qquad
    \subfigure[Feature space dimension fixed to 5, x-axis is number of training samples.]{\includegraphics{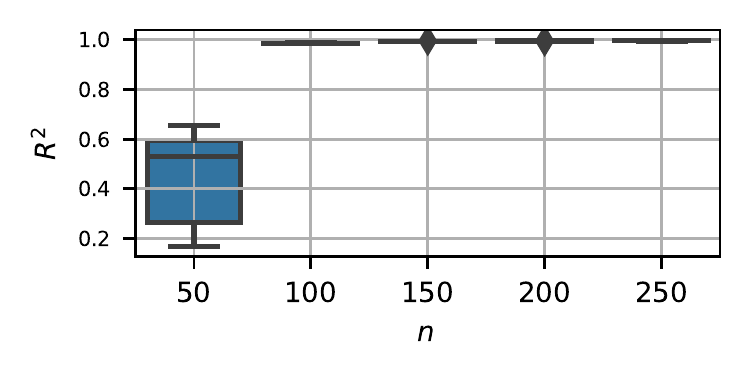}}\\
    \caption{Values of the coefficient of determination $R^2$ as a function of the feature space dimension $m$ for two fixed numbers of training samples (a, b), and as a function of the number $n$ of training samples for two fixed feature space dimensions (c, d).}
    \label{fig:r_squared}
\end{figure}

\begin{figure}%
    \centering
    \subfigure[Number of training samples fixed to 100, x-axis is number of feature space dimensions.]{\includegraphics{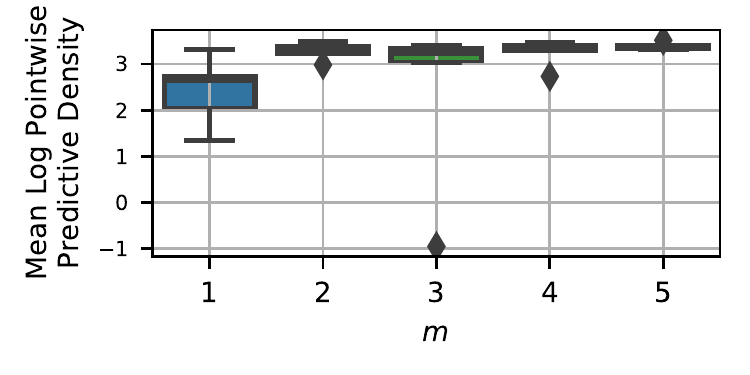}}\qquad
    \subfigure[Number of training samples fixed to 250, x-axis is number of feature space dimensions.]{\includegraphics{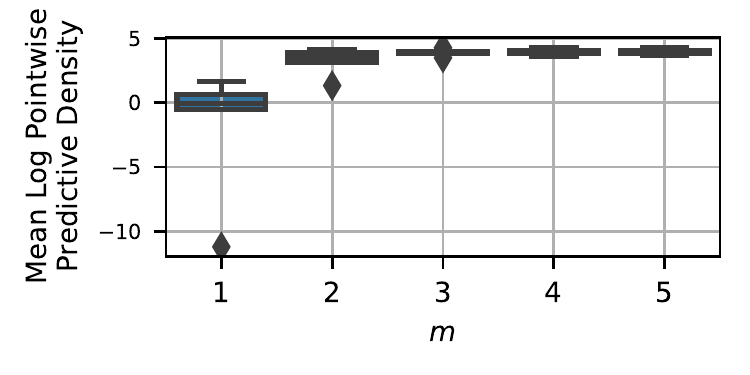}}\\
    \subfigure[Feature space dimension fixed to 1, x-axis is number of training samples.]{\includegraphics{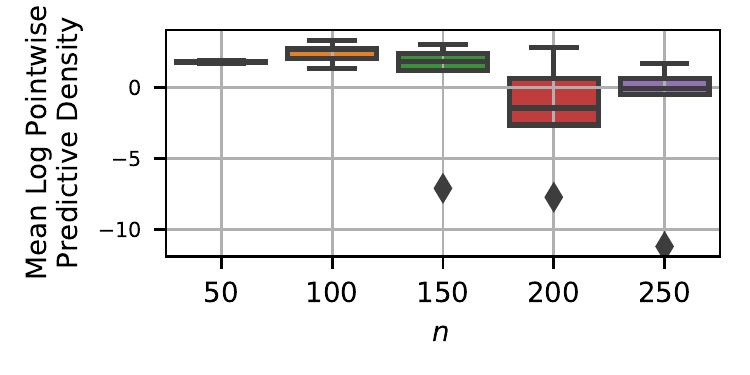}}\qquad
    \subfigure[Feature space dimension fixed to 5, x-axis is number of training samples.]{\includegraphics{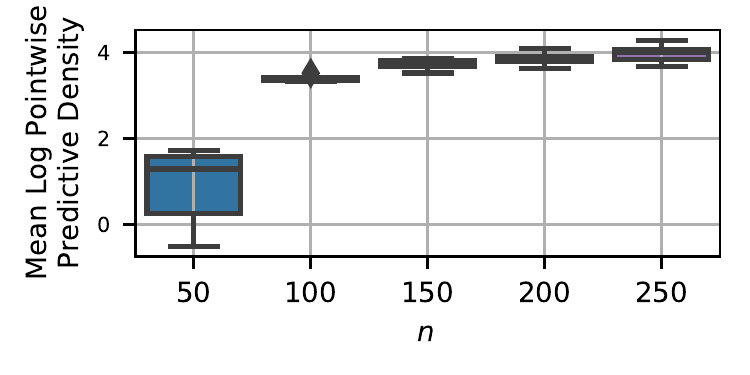}}\\
    \caption{Values of the mean log pointwise predictive density as a function of the feature space dimension $m$ for two fixed numbers of training samples (a, b), and as a function of the number $n$ of training samples for two fixed feature space dimensions (c, d).}
    \label{fig:mlppd}
\end{figure}

\rev{The results presented for fixed numbers of \gls{FS} dimension and number of training samples have been produced with 15,000 draws-long \gls{MCMC} chains preceded by after a 5,000 draws-long warmup phase. These correspond to the results presented in figures \ref{fig:mcmc_chains}, \ref{fig:prior_vs_posterior}, \ref{fig:training_actual_vs_predicted}, and \ref{fig:validation_actual_vs_predicted}. Due to the high computational cost incurred during model inference, shorter 1,000 draws-long chains preceded by a 500 draws-long warmup phase are used for parametric studies, such as those shown in figures \ref{fig:split_gelman_rubin}, \ref{fig:training_duration}, \ref{fig:r_squared}, and \ref{fig:mlppd}.}

\rev{\subsubsection{Statistics Pertaining to Model Training}}
\rev{In this section, we are assessing the validity of the training approach detailed in section \ref{sec:fully_bayesian_as}, that consists in inferring posterior distributions for all model parameters (parameters of the projection parameters as well as \gls{GP} hyperparameters) using \gls{MCMC}.
The quality of the \gls{MCMC} inference process can be assessed by observing the posterior chains (figure \ref{fig:mcmc_chains}) as well as \gls{MCMC}-specific statistics, such as the split Gelman-Rubin statistic (figure \ref{fig:split_gelman_rubin}) \cite{Gelman1992InferenceSequences}.}

\rev{The \gls{MCMC} chains are displayed in figure \ref{fig:mcmc_chains} for two different \gls{FS} dimensions (1D and 5D) and two different number of training samples (100 and 250). 
Only the first five projection parameters $\theta_{p,i}$ are shown, as it would be impractical to show all 50 (1D case) or 240 (5D case) of them. 
We observe that increasing the \gls{FS} dimension and the number of training samples both lead to poor mixing of the chains. 
While adequate mixing occurs in the 1D/100 samples cases for all parameters, increasing the number of training samples leads to poor mixing of the projection parameters, while the quality of the \gls{GP} hyperparameters chains remains satisfactory. 
When increasing the number of \gls{FS} dimensions, the quality of all chains deteriorates. 
An increase of the \gls{FS} dimension leads to an increase of the number of projection parameters, which may explain the poor mixing observed when the \gls{FS} dimension increases may be attributed.
An increase of the number of observations leads to a more sharply peaked posterior distribution, which is accordingly more challenging to sample, and may explain the poor mixing observed as the number of training samples is increased.}

\rev{The evolution of the split Gelman-Rubin statistic shown in figure \ref{fig:split_gelman_rubin} confirms previous observations made on the \gls{MCMC} chains. Both the number of \gls{FS} dimensions and the number of observations lead to an increase in the value of the statistic, indicating a poor approximation of the posterior distribution.
We have seen the \gls{MCMC} inference is made increasingly harder as both the number of training samples and the \gls{FS} dimension increase.}

\rev{Figure \ref{fig:prior_vs_posterior} shows a comparison of the prior and posterior parameter distributions. We recall that the prior distributions were given in algorithm \ref{alg:fully_bayesian_model}. As before, plots have been restricted to the first five projection parameters. While projection parameter distributions are closely concentrated around well-defined values in the 1D case, we observe that they are spread in the 5D case.}

\rev{Figure \ref{fig:training_actual_vs_predicted} depicts a comparison of the actual and predicted lift values for the observations used to train the model. We can see that the model successfully fits the data it was provided, despite the poor mixing observed in the \gls{MCMC} chains.}

\rev{Figure \ref{fig:training_duration} shows evolution of the training duration as a function of the feature space dimension and number of training samples. As expected since the model relies on \gls{GPR}, training duration increases as a power-law as a function of the number of training samples. The impact of the feature space dimension is mostly visible as it increased from 1 to 2, after which no clear upward or downward trend can be observed.}

\rev{We can see that despite poor \gls{MCMC} mixing, the model successfully fits the training data and there is effective learning, as shown by shift from the prior to the posterior model parameter distributions. In the next section, we will assess the predictive performance of the model.}

\rev{\subsubsection{Predictive Performance}}

\rev{Figure \ref{fig:validation_actual_vs_predicted} depicts the comparison of the actual and predicted lift values for the validation points. We observe that the model satisfactorily generalized to the prediction of points that were not used to train the model, demonstrating the utility of the proposed approach to generate a surrogate model. While figure \ref{fig:validation_actual_vs_predicted} does not allow to make clear-cut observations pertaining to the effect of the feature space dimension or the number of training samples, we study the impact of these factors on global predictive accuracy metrics in the next two figures.}

\rev{Figure \ref{fig:r_squared} shows the evolution of the coefficient of determination as a function of both the number of feature space dimensions and training samples. An interesting observation can be made from the comparison of a) and b), that respectively correspond to low and high numbers of training samples. In the former case, increasing the \gls{FS} dimension does not have a significant impact on the model's predictive accuracy while it does in the latter case. When only sparse observations are available, a single direction may then be sufficient to capture the observed variability while more dimensions become necessary as more numerous, and diverse, observations become available. Figures c) and d) show that irrespective of the number of \gls{FS} dimensions, the number of training samples has the expected positive impact on the model's predictive accuracy.}

\rev{Figure \ref{fig:mlppd} shows the evolution of the mean log pointwise predictive density as a function of both the number of feature space dimensions and training samples. Trends are slightly different from those observed with the coefficient of determination. Here, irrespective of the number of training samples, the \gls{MLPPD} always tends to increase with the number of \gls{FS} dimensions. While the \gls{MLPPD} does increase with the number of training samples when the \gls{FS} dimension is 5, it tends to decrease for a single-dimensional \gls{FS}, indicating that the quality of probabilistic predictions is reduced.}
\subsection{Comparative Study}
\label{sec:comparative_study}
A comparative study is carried out to characterize the performance of the proposed method.
\rev{This section starts with a presentation of the benchmark methods used for comparison.}
The next two sections present results for each group of datasets: analytical functions \rev{first} and datasets from science and engineering \rev{afterwards}. 
For both groups, results on all four metrics of interest are presented and discussed.
\subsubsection{Benchmark Methods}
\rev{The two state-of-the-art methods used as benchmarks are presented thereafter. 
We recall that the proposed fully Bayesian method discussed in section~\ref{sec:fully_bayesian_as} is referred to as \glsxtrfull{BAS}}.
\paragraph{MO-AS}
\rev{The first benchmark method was proposed in~\cite{Rajaram2020Non-IntrusiveSubspace} and is based on the \emph{Gaussian processes with built-in dimensionality reduction} method proposed in \cite{Tripathy2016} and discussed in section \ref{sec:gradient_free_as}.
The benchmark method from~\cite{Rajaram2020Non-IntrusiveSubspace} modified two aspects of the method in~\cite{Tripathy2016}: a) a state-of-the-art manifold optimization library~\cite{Townsend2016Pymanopt:Differentiation} was employed, and b) optimization is performed in the Grassmann manifold instead of the Stiefel manifold. 
Here, we only retain the first modification (the state-of-the-art manifold optimization algorithm) but use the Stiefel manifold instead of the Grassmann manifold for the same reasons that we use the Stiefel manifold in our proposed fully Bayesian approach (see discussion in section \ref{sec:fully_bayesian_as}): the directional information retained when working in the Stiefel manifold, which matters for the GP's ARD kernel, is lost when working in the Grassmann manifold. 
This benchmark method is referred to as \gls{MOAS} in the rest of this section. 
In the same spirit as our proposed method, it aims at simultaneously identifying an orthogonal projection onto a low-dimensional input subspace and the mapping from this subspace to the output space. 
However, it does not include any mechanism for quantifying uncertainty in the identified subspace.}
\paragraph{B-GP}
The second benchmark method is based on the method proposed in~\cite{Wycoff2019SequentialSubspaces} and \rev{we refer to it} as \gls{BGP} in this section. 
In this method, a \gls{GPR} model is first built on the full-dimensional input space. 
Instead of gathering gradient samples and using an \gls{MC} approximation \rev{as described in section \ref{sec:active_subspace}}, \rev{the matrix $\mat{C}$ can be analytically derived based on the assumptions of the \gls{GPR} model.} 
\rev{We recall from section \ref{sec:active_subspace} that the \gls{AS} can be obtained by performing an \gls{SVD} of this matrix.
If Bayesian inference is used to train the \gls{GPR} model and posterior distributions for its hyperparameters are obtained, uncertainty can then be propagated using \gls{MC} to obtain a distribution on the \gls{AS}.} 

The computation of the \rev{matrix $\mat{C}$} implemented in the context of this study is only semi-analytical. \gls{GP} gradient evaluations are made analytically once the posterior distribution of the GP hyperparameters has been inferred, but \rev{an \gls{MC} estimator is used to approximate $\mat{C}$ with 1000 gradient samples instead of the fully analytical scheme from~\cite{Wycoff2019SequentialSubspaces}}. 
This was done because an exact reproduction of the process in \cite{Wycoff2019SequentialSubspaces} brought excessive complexity and long runtimes. 
Given the high number of gradient samples used for the \gls{MC} approximation, this modification is not expected to alter results. 
\rev{Improving on the methodology proposed in the original paper, we carry out exact inference of the \gls{GP} hyperparameters distribution using \gls{MCMC} instead of approximate inference.}
\paragraph{Implementation Details}
The \gls{BAS} and \gls{BGP} methods were implemented using the probabilistic programming language \emph{numpyro} \cite{Bingham2019Pyro:Programming, Phan2019ComposableNumPyro}, which uses \emph{JAX} \cite{Bradbury2018JAX:Programs} as computational backend. The \gls{MOAS} method was implemented using a version of the \emph{Pymanopt} framework \cite{Townsend2016Pymanopt:Differentiation} modified to use \emph{JAX}. Using the same computational backend across all three methods helps in ensuring a level playing field for the comparative study such that differences in training time can be linked to the methods themselves instead of implementation specifics. For the Bayesian methods, MCMC is used for inference, leveraging the No U-Turn Sampler (NUTS) as implemented in \emph{numpyro}. Four parallel chains are sampled, each consisting of 1,000 samples from the joint posterior probability distribution on model parameters and initialized using 500 warmup samples. For \gls{MOAS}, 500 restarts of the manifold optimization algorithm are used, as in \cite{Rajaram2020Non-IntrusiveSubspace}. The implementation of the three models used to produce the results presented in this paper are available online\footnote{\url{https://gitlab.com/raphaelgautier/bayesian-supervised-dimension-reduction}}.
\subsubsection{Results on Quadratic Functions}
\paragraph{Active Subspace Recovery}
\begin{figure}[p]
    \centering
    \includegraphics{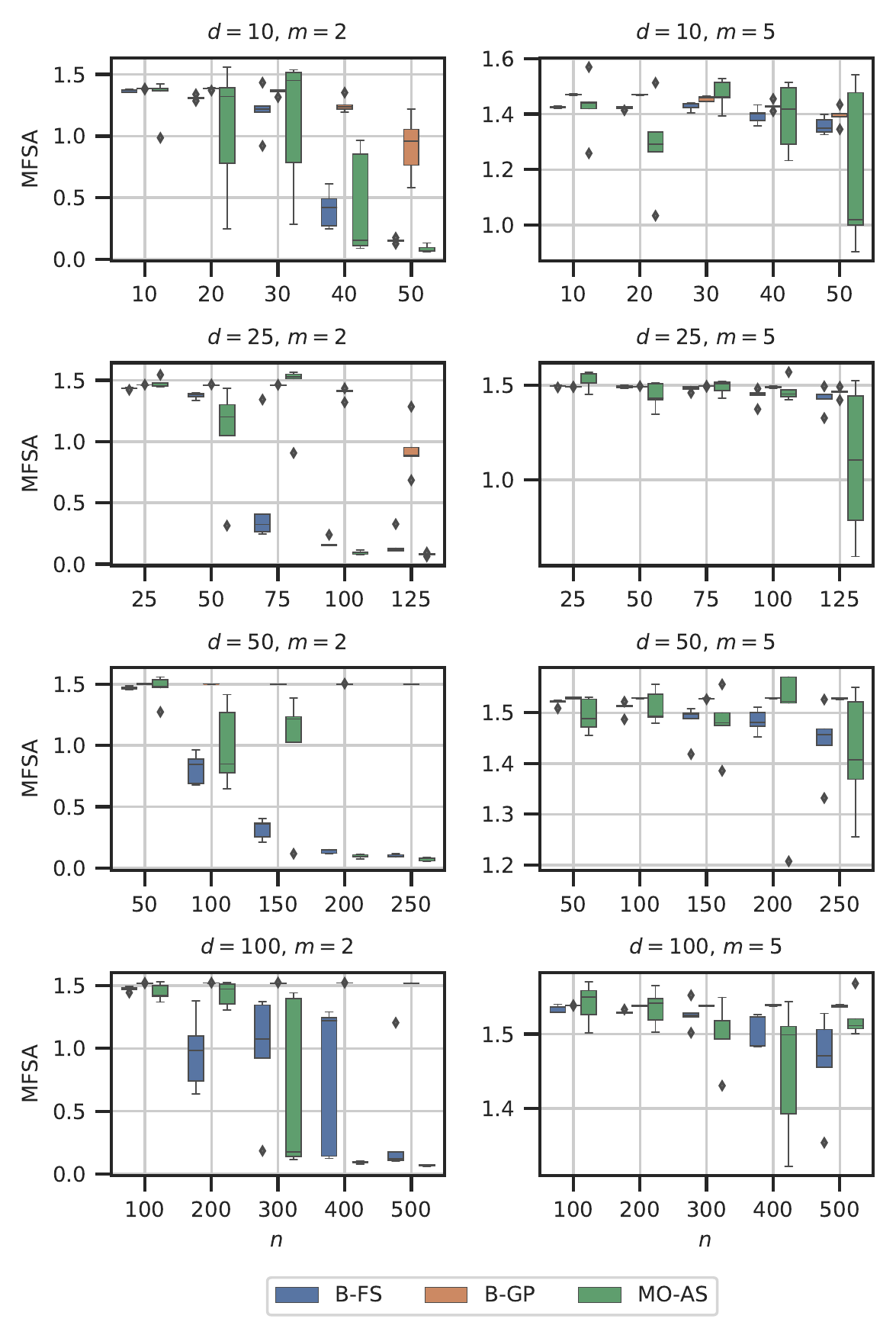}
    \caption{Evolution of the \glsxtrfull{MFSA} between the predicted and actual active subspaces for training sets varying in size $n$ from one to five times the number of input dimensions for quadratic function datasets. Plots are organized by increasing input dimension $d$ from top to bottom and by increasing AS dimension $m$ from left to right.}
    \label{fig:quadratic_functions_subspace_angle}
\end{figure}

Figure \ref{fig:quadratic_functions_subspace_angle} depicts the evolution of the \rev{\gls{MFSA}} as a function of the number of samples used to train the model. A low subspace angle indicates that the \rev{uncovered feature space and the \gls{AS}} are nearly aligned, corresponding to a successful recovery of the \gls{AS}.

\rev{For the cases featuring a 2D feature space ($m=2$)}, trends seem to indicate that a minimum number of training samples is required to successfully recover the AS, which is indicated by the drop of the subspace angle metric.
For all eight benchmark quadratic functions, we observe that \gls{BGP} consistently exhibits worse \gls{AS} recovery capabilities than the other two methods. 
The performance of \gls{BGP} also decreases with the number of input dimensions.
While the drop in first subspace angle, characteristic of successful AS recovery, is indeed visible within the range of training set sizes under study when $d=10$ and $d=25$, no such drop is visible when $d=50$ and $d=100$. 
Actively seeking the \gls{AS} by adapting the form of the predictive model to incorporate a projection onto a lower-dimensional subspace therefore appears to drastically improve the ability of surrogate-based methods to detect the \gls{AS} when a limited number of training samples is available.
Both the \gls{BAS} and \gls{MOAS} have similar \gls{AS} recovery capabilities. As noted earlier, the distributions for the \gls{BAS} method have wider spread due to the Bayesian nature of the method: instead of only seeking the most likely \gls{AS}, the proposed \gls{BAS} gives access to the full posterior distribution of the \gls{AS}. Directions that are less likely given the model observations are therefore retained and given a smaller weight. We will see that it enables better quantification of the epistemic uncertainty. We also note the skewness of some of those distributions, such as for $d=25$, $m=1$, and 50 or 75 training samples: while the distribution almost spans the complete interval of subspace angle values, most of its weight is concentrated on small angle values. Both methods consistently enable the detection of the \gls{AS} a number of training samples smaller than five times the number of dimensions for 25- to 100-dimensional quadratic functions.

\rev{For the cases featuring a 5D feature space ($m=5$), we always observe high subspace angle values which indicate that the feature space and \gls{AS} are misaligned in at least one direction. As we will see however, this does not necessarily significantly impact the predictive accuracy of the model. This may be explained by the fact that even though not all \gls{AS} directions have been identified, a subset of them may have actually been properly recovered.} 
\paragraph{Deterministic Predictive Capability}
\begin{figure}[p]
    \centering
    \includegraphics{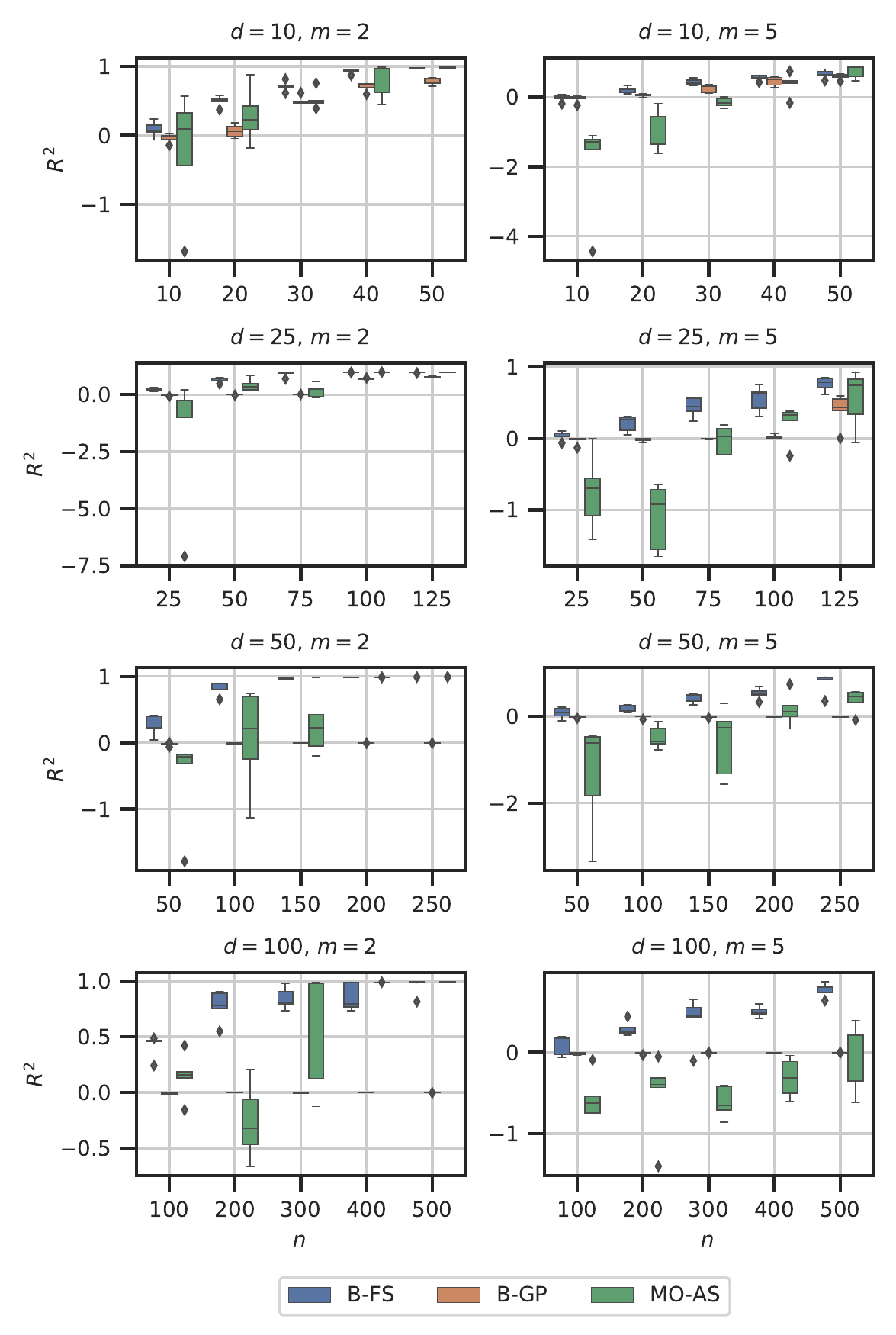}
    \caption{Evolution of the validation \glsxtrfull{r2} for training sets varying in size $n$ from one to five times the number of input dimensions for quadratic function datasets. Plots are organized by increasing input dimension $d$ from top to bottom and by increasing AS dimension $m$ from left to right.}
    \label{fig:quadratic_functions_rmse}
\end{figure}
Figure \ref{fig:quadratic_functions_rmse} depicts the evolution of the \glsxtrlong{r2} as a function of the number of samples used to train the predictive model. For both Bayesian approaches, the median is used for point-based prediction.

The deterministic predictive capability of \gls{BGP} for high-dimensional input spaces is consistent with its poor ability to detect the \gls{AS}: it nearly always scores worse than both other methods. The relative drop in performance with an increasing number of input dimensions first observed with the first subspace angle is here confirmed: \rev{when $m=2$}, the deterministic predictions of \gls{BGP} do not improve over the studied range of training samples neither for $d=50$ nor for $d=100$.

The comparison of the \gls{BAS} and \gls{MOAS} methods indicates that both methods perform equally well for a number of training samples ranging from three to five times the number of input space dimensions \rev{when $m=2$}. For very small training sets \rev{and for a higher-dimensional feature space ($m=5$)} however, \rev{\gls{BAS}} invariably displays much higher \gls{r2} values than the other two methods. This is consistent with the expected superiority of Bayesian methods when few model observations are available. In the light of those results, the proposed \gls{BAS} approach therefore appears to improve upon the \gls{MOAS} approach: \rev{at worst it reaches the same performance as \gls{MOAS} when a relatively high number of training samples are available and the feature space dimension is low, and it significantly increases predictive performance in the sparse data regime and for higher-dimensional feature spaces.}
\paragraph{Probabilistic Predictive Capability}
\begin{figure}[p]
    \centering
    \includegraphics{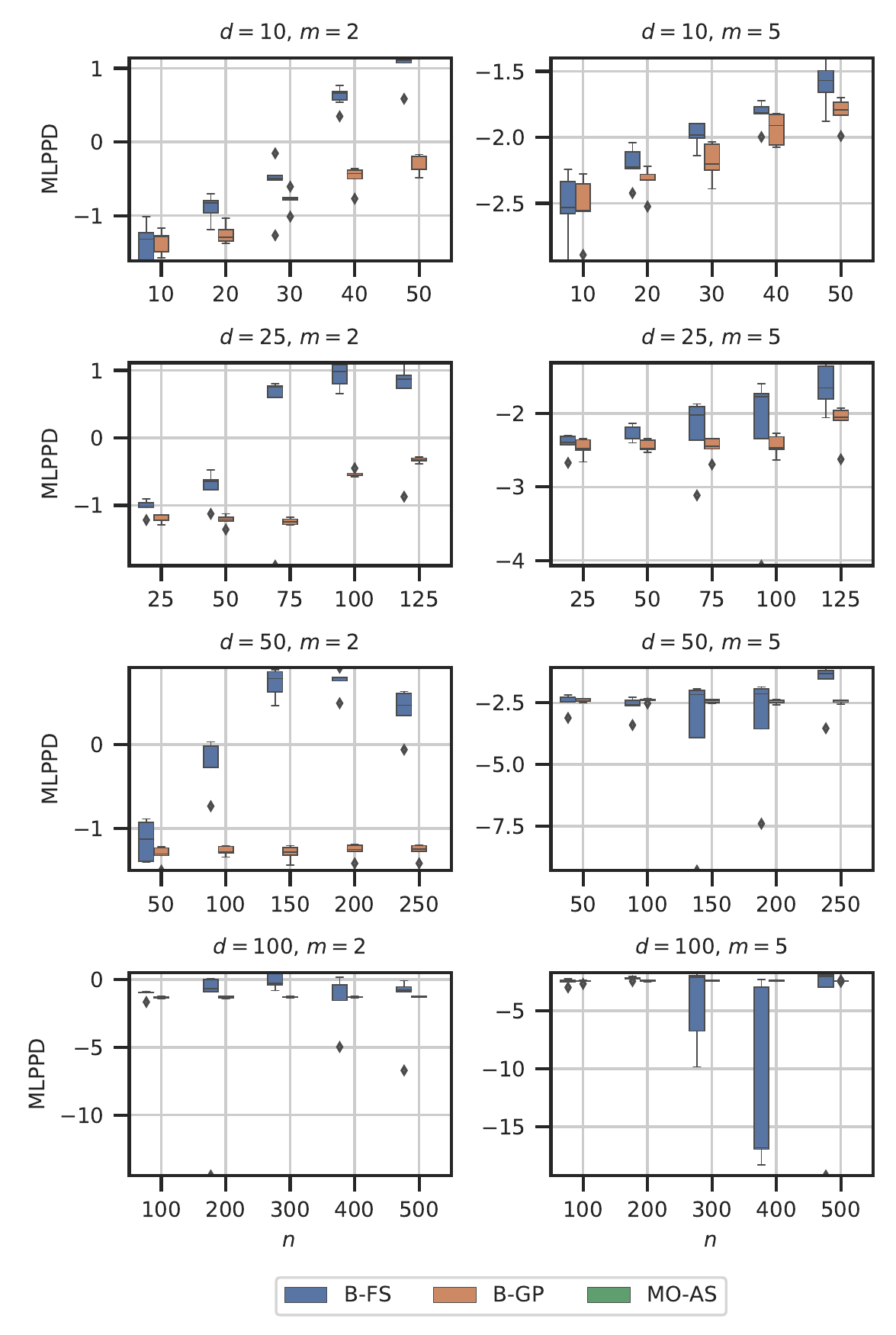}
    \caption{Evolution of the \glsentryfull{MLPPD} for training sets varying in size $n$ from one to five times the number of input dimensions for quadratic function datasets. Plots are organized by increasing input dimension $d$ from top to bottom and by increasing AS dimension $m$ from left to right.}
    \label{fig:quadratic_functions_log_likelihood}
\end{figure}
Figure \ref{fig:quadratic_functions_log_likelihood} shows the evolution of the \gls{MLPPD} as the number of training samples is increased. This metric reveals the quality of the probabilistic prediction: the higher the \gls{MLPPD}, the most likely it is to observe the actual responses of the validation dataset under the posterior of the predictive model. 

As expected, both Bayesian methods consistently score better than \gls{MOAS} with respect to this metric. 
\rev{\gls{MOAS} actually scores so low that those values were filtered out of the plots to enable readability.} 
This can be explained by the fact that uncertainty in the \gls{MOAS} model is only partially captured: while the GP effectively captures part of the uncertainty, neither the uncertainty in the GP hyperparameters nor in the projection matrix parameters \rev{are} quantified. 
\rev{Except for the most challenging cases ($d=50,m=5$ and $d=100,m=5$), the proposed \gls{BAS} method generally leads to the highest \gls{MLPPD} values.}
\rev{Even though \gls{BGP} scores better in terms of \gls{MLPPD} for $d=50,m=5$ and $d=100,m=5$, we recall that \gls{r2} was nearly zero for \gls{BGP} in these cases, which indicated very poor point-based predictive accuracy. Conclusions can therefore not be drawn regarding the probabilistic predictive accuracy of the two methods using \gls{MLPPD} in these cases.} 

\paragraph{Training Duration}
\begin{figure}[p]
    \centering
    \includegraphics{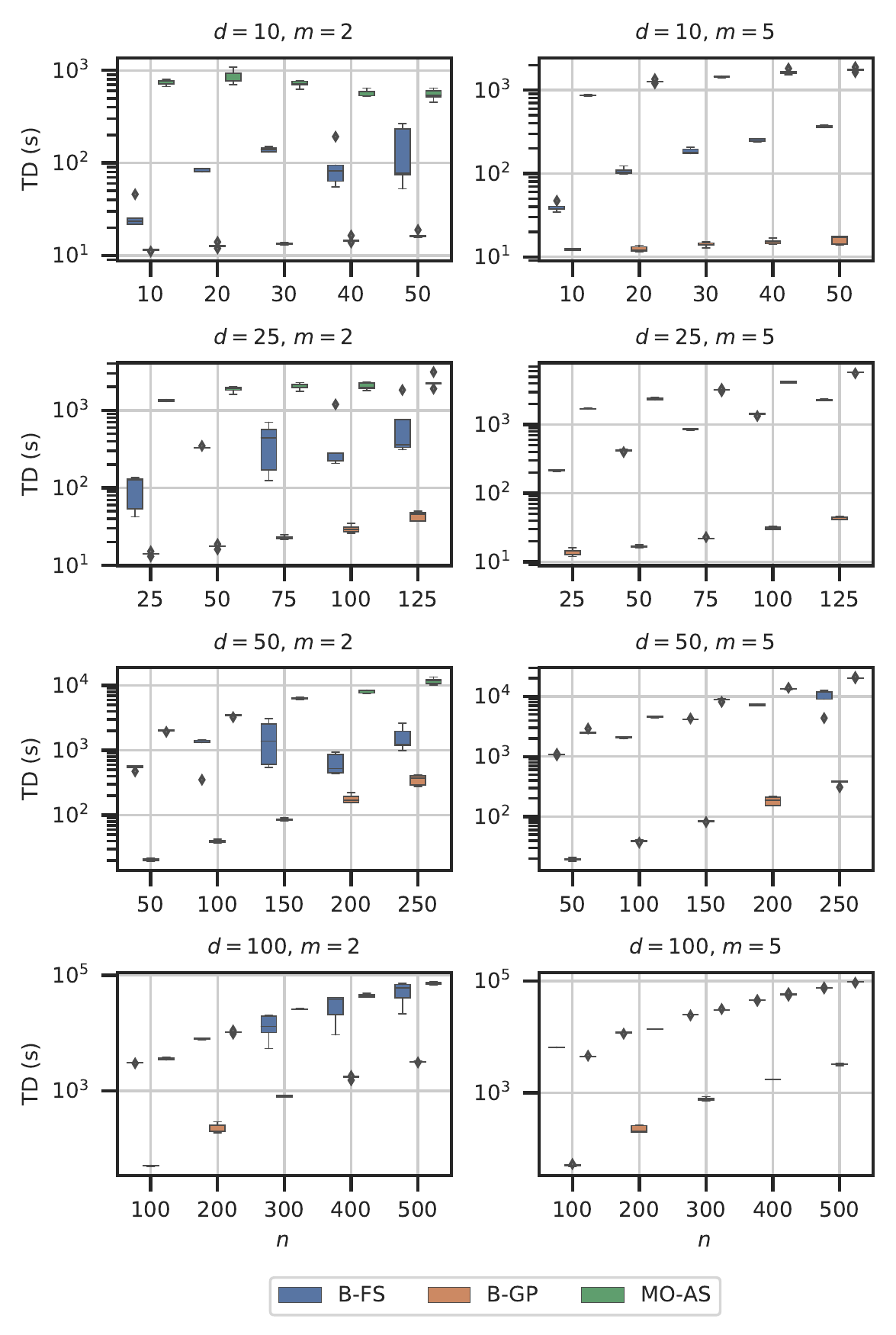}
    \caption{Evolution of the training time (TT) for training sets varying in size $n$ from one to five times the number of input dimensions for quadratic function datasets. Plots are organized by increasing input dimension $d$ from top to bottom and by increasing AS dimension $m$ from left to right.}
    \label{fig:quadratic_functions_training_time}
\end{figure}
Figure \ref{fig:quadratic_functions_training_time} depicts the evolution of the time required to train all three types of models for the different quadratic functions under study. Training time always increases with the number of training samples. This is expected as the cost of computing the inverse of the sample covariance matrix during the \gls{GP} likelihood computation increases as the matrix size increases. For \rev{most figures}, a linear trend in log-scale is clearly visible, that corresponds to a power-law scaling in linear scale. Such a behavior is expected for \gls{BGP}, for which the $\mathcal{O}(n^3)$ scaling is well-known, where $n$ is the number of training samples. 

The \gls{BGP} models are consistently faster to train than both projection-based methods. This is expected since those methods require additional computations during \rev{the evaluation of the model likelihood} and increase the number of parameters to be inferred or optimized without affecting the size of the sample covariance matrix, which is the bottleneck of \gls{GP} training. 

While the proposed \gls{BAS} model trains faster than the \gls{MOAS} model for low input space dimensions and low number of training samples, its training duration becomes larger when dimension and training samples increase. \rev{At worst, \gls{BAS} and \gls{MOAS} training times are of the same order.} This may be explained by considering the differences between the two methods. On the one hand, in \gls{MOAS}, the number of restarts of the gradient-based optimization is fixed but the number of steps during one optimization run (and therefore the number of likelihood evaluations) may vary because a dynamic stopping criterion is used to terminate optimization. On the other hand, in \gls{BAS}, the \gls{MCMC} chain length is fixed but the size of the leapfrog steps of the \gls{NUTS} sampler are adaptively chosen during warmup based on the shape of the likelihood function. In practice, we observed greater variability in the \gls{MCMC} step size than we did in the number of optimization steps. As the number of input dimensions increases and the likelihood function becomes more challenging to sample, the step size chosen by the \gls{NUTS} algorithm decreases, thus leading to more \gls{HMC} leapfrog steps and increased total sampling times.
\subsubsection{Results on Science and Engineering Datasets}
This section mirrors the preceding section by presenting results for all four metrics of interest, this time on the benchmark science and engineering datasets. While the quadratic functions may be representative of simple functions encountered in practical engineering applications, these datasets were generated using actual analyses encountered in scientific or engineering practice. 
\paragraph{Active Subspace Recovery}
\begin{figure}[ht]
    \centering
    \includegraphics{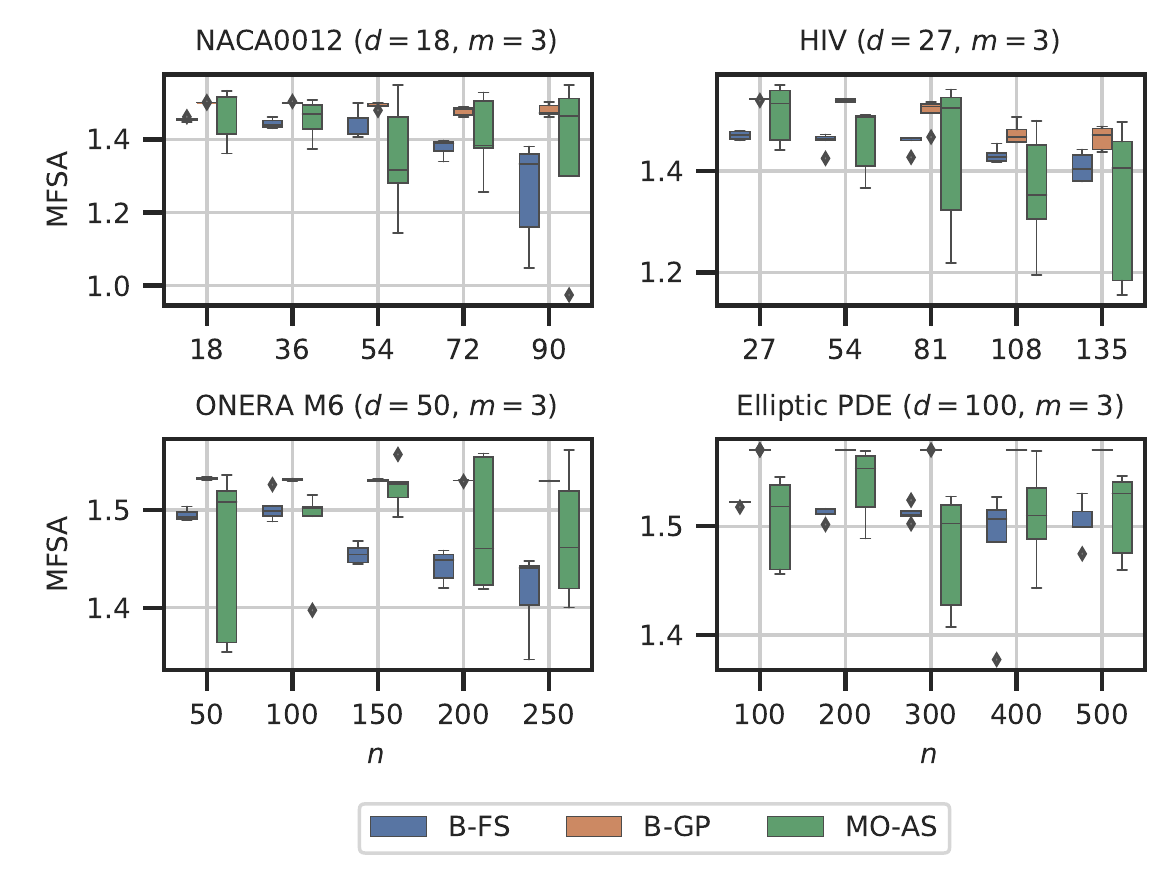}
    \caption{Evolution of the first subspace angle (FSA) between the predicted and actual active subspaces for training sets varying in size $n$ from one to five times the number of input dimensions for all four science and engineering datasets.}
    \label{fig:real_datasets_subspace_angle}
\end{figure}
Figure \ref{fig:real_datasets_subspace_angle} presents the \gls{AS} recovery results obtained for science and engineering datasets. \rev{We observe that the recovered feature spaces never correspond to the true \gls{AS}. As noted before, we will again see that this does not necessarily translate into a poor predictive accuracy.}
\paragraph{Deterministic Predictive Capability}
\begin{figure}[ht]
    \centering
    \includegraphics{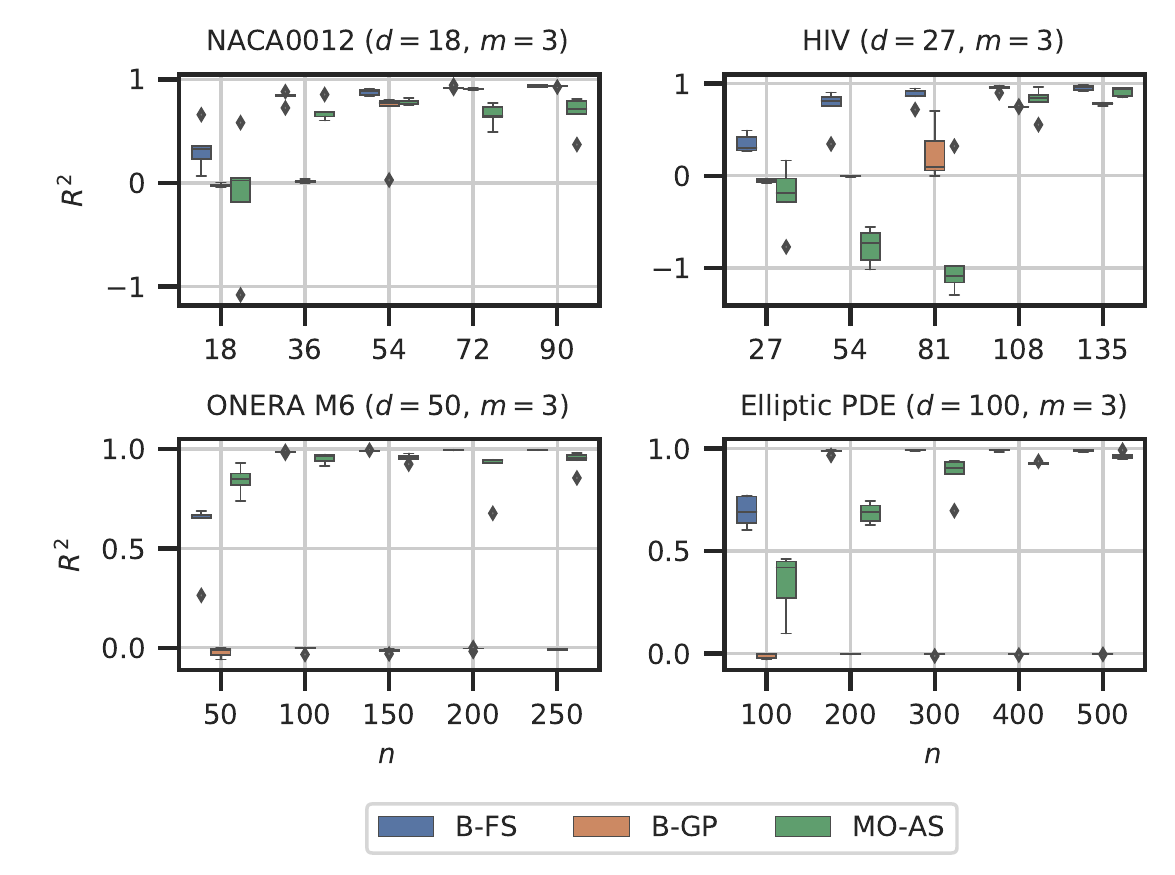}
    \caption{Evolution of the validation \glsxtrfull{r2} for training sets varying in size $n$ from one to five times the number of input dimensions for all four science and engineering datasets.}
    \label{fig:real_datasets_rmse}
\end{figure}
Figure \ref{fig:real_datasets_rmse} shows the evolution of the \glsxtrlong{r2} as a function of the number of training samples for the science and engineering datasets. \gls{BAS} outperforms both other methods on all four datasets. Where \gls{BGP} fails to produce a useful model for any number of training samples within the domain of study for both the ONERA M6 and Elliptic PDE datasets, \gls{BAS} yields well-performing models with as little as a number of training samples corresponding to twice the number of input dimensions. For the NACA0012 and HIV datasets where \gls{BGP} eventually leads to satisfactory models, the proposed \gls{BAS} approach gives access to better models with significantly fewer training samples.
\paragraph{Probabilistic Predictive Capability}
\begin{figure}[ht]
    \centering
    \includegraphics{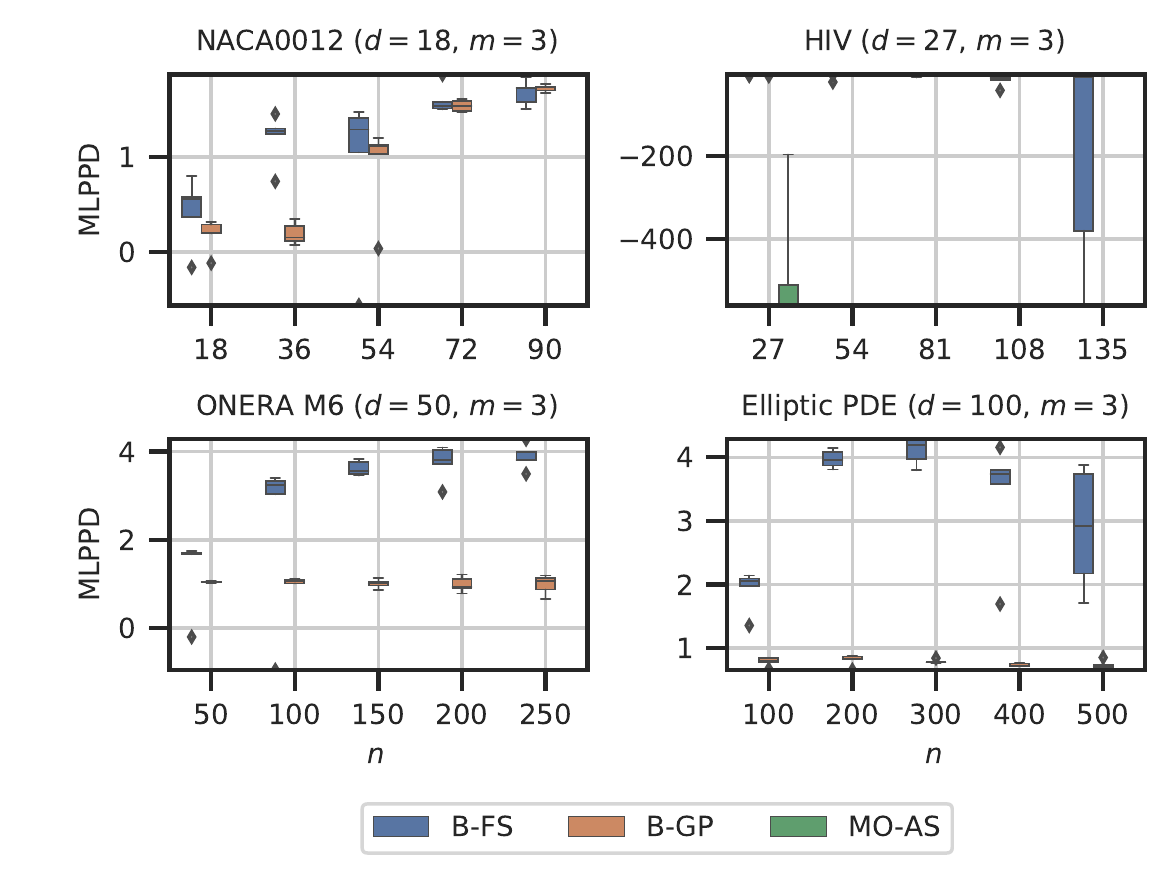}
    \caption{Evolution of the \glsxtrfull{MLPPD} for training sets varying in size $n$ from one to five times the number of input dimensions for all four science and engineering datasets.}
    \label{fig:real_datasets_log_likelihood}
\end{figure}

\rev{Figure \ref{fig:real_datasets_log_likelihood} depicts the evolution of the validation \gls{MLPPD} for different numbers of training samples. Those graphs have been truncated because the \gls{MLPPD} values for the \gls{MOAS} method are consistently significantly lower than the other two methods and compromise the readability of those graphs.} As opposed to the results shown in \ref{fig:quadratic_functions_log_likelihood} that were mostly consistent across different numbers of input space and active subspace dimensions, the corresponding results for science and engineering datasets seem to be highly problem-dependent.

For the relatively low-dimensional NACA0012 dataset, \gls{MOAS} displays the poorest probabilistic predictive capabilities. \gls{BAS} exhibits better performance than \gls{BGP} for small training sets, but is slightly outperformed by \gls{BGP} for larger training sets. This is an illustration of the trade-off occurring when reducing the dimensionality of the inputs. While \gls{BGP} operates on the full input space, the proposed method always operates on one of its low-dimensional subspaces. As a result, as the number of training samples increases, \gls{BGP} can start capturing variability in the response due to variations of the inputs in the inactive subspace while \gls{BAS} is limited to capturing variations of the response due to variations of the inputs in the active subspace only, and variations in the inactive subspace are captured as noise.

\rev{The 27-dimensional HIV dataset appears to be particularly challenging for both the \gls{BAS} and \gls{MOAS} methods, as extremely low values of \gls{MLPPD} are reached. This behavior seems to be highly problem-dependent as it is not encountered for any other of the datasets.}

The performance of the proposed \gls{BAS} method significantly improves for the two highest-dimensional datasets, ONERA M6 and Elliptic PDE, with consistently higher \gls{MLPPD} than both other methods across all training set sizes.

\paragraph{Training Duration}
\begin{figure}[ht]
    \centering
    \includegraphics{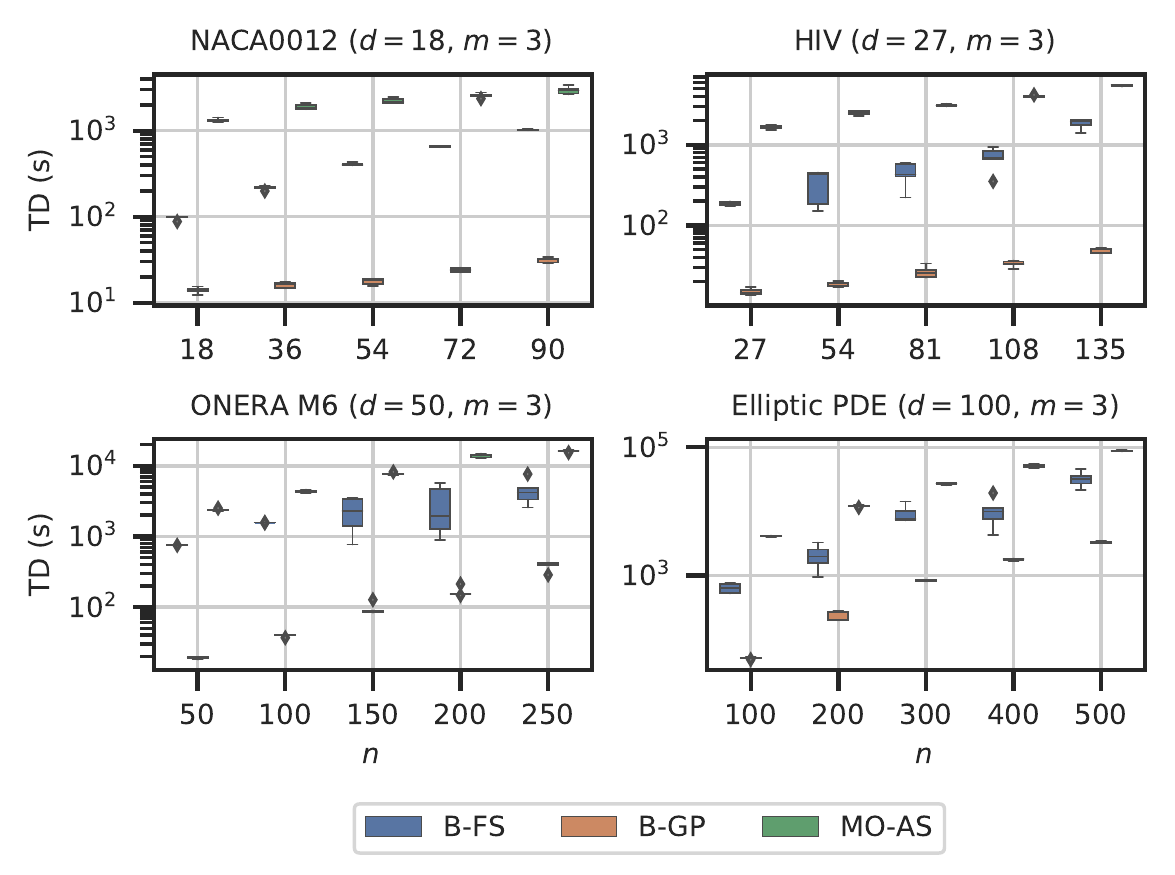}
    \caption{Evolution of the training time (TT) for training sets varying in size $n$ from one to five times the number of input dimensions for all four science and engineering datasets.}
    \label{fig:real_datasets_training_time}
\end{figure}
Figure \ref{fig:real_datasets_training_time} shows the evolution of the training time as a function of the number of training samples. \gls{BGP} models are consistently faster to train, as fewer model parameters need to be identified compared to projection-based methods. On those four datasets, \rev{the training duration is consistently inferior for \gls{BAS} than it is for \gls{MOAS}}, even in the case of the Elliptic PDE featuring a 100-dimensional input space. Beyond the expected scaling with the number of training samples and the number of inference parameters, training time appears to be highly dependent on the problem-at-hand.

\section{Conclusion}
\label{sec:conclusion}
The proposed \gls{BAS} method was designed to assist the creation of surrogate models of computationally expensive analyses with high-dimensional input spaces and for which access to gradients is not available. It enriches the family of projection-based methods for supervised dimension reduction with a gradient-free and fully Bayesian alternative. 
This work offered a comparative study of the proposed method with two other state-of-the art methods, \gls{MOAS} and \gls{BGP}, that focused on four aspects: recovery of the active subspace, deterministic prediction accuracy, probabilistic prediction accuracy, and training time.

The study was carried out on eight analytical functions (25 to 100 inputs) and four science and engineering datasets (18 to 100 inputs) and showed the proposed method to be superior to previously introduced methods mainly due to its improved probabilistic predictive ability. 
Where optimization-based methods confidently make wrong predictions, the proposed method adequately estimates the uncertainty in its predictions thanks to the fully Bayesian approach. 
The explicit incorporation of a projection onto a lower-dimensional subspace within the form of the surrogate model was shown to ease the identification of the \gls{AS} and in turn to improve the predictive capabilities of the resulting surrogate model, as opposed to the surrogate-based approaches to \gls{AS}, such that \gls{BGP}, that aim at first constructing a full-dimensional surrogate model and then using it to find the \gls{AS}.

While the dimension of the \gls{AS} was assumed to be known throughout the present study, this is not the case in practice when being confronted to a new dataset. 
Existing methods to assess the \gls{AS} dimension have been proposed, notably alongside the two benchmark methods used in this study. 
In \cite{Tripathy2016}, the authors propose to successively train the model assuming different \gls{AS} dimensions and select the dimension based on the \gls{BIC}. 
This method may be deemed unsatisfactory as it requires multiple costly training runs to obtain a single model. 
In \cite{Wycoff2019SequentialSubspaces}, the original \gls{AS} methodology for selecting the number of active dimensions can be carried out since a surrogate model in the full-dimensional input space is built. 
However, as shown in this study, significantly more samples are required to obtain a good model in the full-dimensional input space compared to methods explicitly incorporating the projection onto a low-dimensional subspace. 
A gradient-free method for determining the \gls{AS} dimension when few model observations are available remains an open problem and will be the subject of future work.

Resorting to a surrogate modeling technique such as the one proposed in this study, that incurs a significant computational cost in addition to the cost of model evaluations, is justified for analyses whose computational cost is itself high. 
Given that realistic computational budgets are limited, high computational cost mechanically results in a small number of model observations. 
Adaptive sampling methods aim at intelligently selecting those few model observations such that the accuracy of the resulting surrogate model is maximized using predictive uncertainty. 
An accurate quantification of predictive uncertainty for relatively small training sets is crucial in that process, as adaptive sampling is carried out when only a limited number of model evaluations are available. 
The proposed method therefore appears as a viable candidate for surrogate-based adaptive sampling, as it exhibits comparable training times and often better probabilistic predictive capabilities than state-of-the-art methods. 
The application of the proposed method within an adaptive sampling scheme tailored to functions with high-dimensional inputs will therefore be another avenue for future work.

\bibliographystyle{unsrt} 
\bibliography{references}

\clearpage

\end{document}